\documentclass[acmsmall,authorversion]{acmart}

\usepackage{graphicx}
\usepackage{natbib}
\usepackage{booktabs}
\usepackage{multirow}
\usepackage{siunitx}
\usepackage{subcaption}
\usepackage{wrapfig}
\usepackage{hyperref}

\usepackage[formats]{listings}
\usepackage[frozencache,cachedir=.]{minted}
\setcopyright{acmlicensed}
\acmJournal{JATS}
\acmYear{2025} \acmVolume{3} \acmNumber{1} \acmArticle{4} \acmMonth{9}
\acmDOI{10.1145/3742440}

\begin{document}

    \title{Hybrid Many-Objective Optimization in Probabilistic Mission Design for Compliant and Effective UAV Routing}

    \author{Simon Kohaut}
    \affiliation{%
      \institution{Artificial Intelligence and Machine Learning Group, TU Darmstadt}
      \city{Darmstadt}
      \country{Germany}
    }
    \email{simon.kohaut@cs.tu-darmstadt.de}
    
    \author{Nikolas Hohmann}
    \affiliation{%
      \institution{Control Methods and Intelligent Systems Group, TU Darmstadt}
      \city{Darmstadt}
      \country{Germany}
    }
    \email{nikolas.hohmann@tu-darmstadt.de}

    \author{Sebastian Brulin}
    \affiliation{%
      \institution{Honda Research Institute EU}
      \city{Offenbach}
      \country{Germany}
    }
    \email{sebastian.brulin@honda-ri.de}
    
    \author{Benedict Flade}
    \affiliation{%
      \institution{Honda Research Institute EU}
      \city{Offenbach}
      \country{Germany}
    }
    \email{benedict.flade@honda-ri.de}
    
    \author{Julian Eggert}
    \affiliation{%
      \institution{Honda Research Institute EU}
      \city{Offenbach}
      \country{Germany}
    }
    \email{julian.eggert@honda-ri.de}
    
    \author{Markus Olhofer}
    \affiliation{%
      \institution{Honda Research Institute EU}
      \city{Offenbach}
      \country{Germany}
    }
    \email{markus.olhofer@honda-ri.de}

    \author{Jürgen Adamy}
    \affiliation{%
      \institution{Control Methods and Intelligent Systems Group, TU Darmstadt}
      \city{Darmstadt}
      \country{Germany}
    }
    \email{juergen.adamy@tu-darmstadt.de}

    \author{Devendra Singh Dhami}
    \affiliation{%
      \institution{Uncertainty in Artificial Intelligence Group, TU Eindhoven}
      \city{Eindhoven}
      \country{Netherlands}
    }
    \email{d.s.dhami@tue.nl}

    \author{Kristian Kersting}
    \affiliation{%
      \institution{Artificial Intelligence and Machine Learning Group, TU Darmstadt}
      \city{Darmstadt}
      \country{Germany}
    }
    \email{kersting@cs.tu-darmstadt.de}


    \begin{abstract}
        Advanced Aerial Mobility encompasses many outstanding applications that promise to revolutionize modern logistics and pave the way for various public services and industry uses.
        However, throughout its history, the development of such systems has been impeded by the complexity of legal restrictions and physical constraints. 
        While airspaces are often tightly shaped by various legal requirements, Unmanned Aerial Vehicles (UAV) must simultaneously consider, among others, energy demands, signal quality, and noise pollution. 
        In this work, we address this challenge by presenting a novel architecture that integrates methods of Probabilistic Mission Design (ProMis)~\cite{kohaut2023mission, kohaut2024ceo} and Many-Objective Optimization~\cite{hohmann2023three} for UAV routing.
        Hereby, our framework facilitates compliance with legal requirements under uncertainty while producing effective paths that minimize various physical costs a UAV needs to consider when traversing human-inhabited spaces. 
        To this end, we combine hybrid probabilistic first-order logic for spatial reasoning with mixed deterministic-stochastic route optimization, incorporating physical objectives such as energy consumption and radio interference with a logical, probabilistic model of legal requirements. 
        We demonstrate the versatility and advantages of our system in a large-scale empirical evaluation over real-world, crowd-sourced data from a map extract from the city of Paris, France, showing how a network of effective and compliant paths can be formed.
    \end{abstract}

    \maketitle
    
    \newpage
\section{Introduction}

Unmanned Aerial Vehicles (UAV) have emerged as a novel form of autonomous transportation systems, with numerous applications in logistics, public, and industry, promising to disrupt the market and revolutionize the respective fields.
However, their employment depends on two factors:
First, the availability of robust routing schemes that can satisfy many physical objectives simultaneously, such as energy and time requirements, signal strengths, or noise pollution in human-inhabited spaces.
Second, the ability to delicately navigate airspace regulations, such as those created for Unmanned Traffic Management (UTM) or adopted by local authorities.
Providing a UAV routing scheme that allows both criteria to be factored in is an outstanding challenge in enabling Advanced Air Mobility (AAM) applications.

With regard to physical objectives, prior work has demonstrated the application of hybrid many-objective optimization techniques~\cite{hohmann2023three}.
That is, by combining exact graph-based search with a metaheuristic evolutionary optimization, a set of smooth paths that individually trade off the pre-defined objectives can be obtained.
Although this enables navigating within physical optimization goals with fine-granular control over their relative importance, modeling and adhering to complex traffic rules remains challenging.
To this end, Probabilistic Mission Design (ProMis) has been presented to encode operational constraints such as UTM rules, manufacturer restrictions, and operator preferences as hybrid probabilistic first-order logic programs~\cite{Kohaut2023}.
Hence, ProMis allows for probabilistic inference across the agent's navigation space, resulting in scalar fields of independent, identically distributed (i.i.d.) probabilities of all modeled laws being satisfied under uncertainty in the environment.
Mission design, as it is crucial for enabling AAM applications such as deciding clearance, explaining the decision, and proposing better mission settings, can be implemented over ProMis~\cite{kohaut2024ceo}.

In this work, we combine these approaches into a single system for UAV routing under uncertainty, allowing AAM scenarios to take place in an effective manner while complying with the local legal restrictions established in human-inhabited spaces through probabilistic and symbolic reasoning.
Given start- and end-points, our framework produces a Pareto set of physically effective and legally compliant routes from which the agent can select according to its individual weighting of their respective importance.
Furthermore, clearance for the resulting paths can be decided independently, allowing one to discard any potential candidates that do not sufficiently respect regulations.
To this end, we demonstrate how the impact of high-level parameters controlling the mission's overall setting, such as the employed license of the pilot or time-of-day, can be explained to the UAV operator and automatically selected for optimal rule adherence.

In summary, we make the following substantial contributions to the field.
\begin{itemize}
    \item We present Probabilistic Mission Design (ProMis) with Hybrid Many-Objective Optimization for UAV routing to jointly consider physical objectives and legal constraints in urban AAM.
    \item We introduce the Constitutional Language (CoLa), a novel description language for encoding an agent's objectives across its state space and paths to combine legal constraints and physical objectives in a unified fashion.
    To this end, CoLa uses spatial reasoning under uncertainty facilitated via Statistical Relational Maps (StaR Maps)~\cite{flade2024star}.
    \item We demonstrate how our system can decide Clearance for a given mission, Explain the impact of high-level semantic parameters, and Optimize for ideal compliance.
    \item We publish our implementations of (i) our Versatile Intelligent Aerial Streets (VIAS\footnote{\href{https://github.com/NikHoh/VIAS}{github.com/NikHoh/VIAS}}) algorithm for 3D path-planning under many objectives and (ii) our overarching system for Probabilistic Mission Design and Routing\footnote{\href{https://www.github.com/HRI-EU/ProMis}{github.com/HRI-EU/ProMis}} as presented in this work.
\end{itemize}

The UAV routing framework presented in this paper builds on and synthesizes several prior works. 
Specifically, we integrate StaR Maps~\cite{flade2024star} as a hybrid probabilistic environment model, and parameterize a probabilistic logic representation of UAV behavior, similar to ProMis~\cite{Kohaut2023}, to enable high-level decision-making during mission execution~\cite{kohaut2024ceo}. 
We expand on these ideas with our CoLa language, which introduces a description layer that unifies social and physical objectives for effective and compliant UAV routing. 
Ultimately, this setup guides the path planning process within our hybrid many-objective optimization framework, VIAS~\cite{hohmann2023three}, allowing us to apply the aforementioned models in a large-scale, 3D routing evaluation employing real-world data.

In the following, we discuss the backgrounds of probabilistic environment representations, many-objective routing, and neuro-symbolic reasoning, and touch upon recent developments of regulatory frameworks for UAVs in Section~\ref{sec:related_work}.

With this overview in mind, we present our methods in Section~\ref{sec:methods}.
Starting with the architecture of our approach, made up of StaR Maps, ProMis, and the VIAS many-objective optimizer, we lay out our framework's building blocks in detail, including the objectives we consider in this work and the syntax and semantics of CoLa.

Our findings are underpinned by an exhaustive experimental study over crowd-sourced, real-world data from Paris, France, in Section~\ref{sec:experiments}.
Hence, we show how our system yields networks of physically effective and legally compliant paths for UAVs, even in complex urban environments.

Finally, we summarize our work, discuss its limitations, and point toward future work in Section~\ref{sec:conclusion}.

    \section{Related Work}
\label{sec:related_work}

When discussing novel approaches to compliance in intelligent transportation systems, it is important to consider the ongoing development of public regulations.
In Europe, the Single European Sky Air Traffic Management Research (SESAR) initiative has significantly influenced the advancement of Unmanned Aerial Vehicle (UAV) applications and Urban Air Mobility (UAM). 
With subsequent editions of the SESAR Master Plan, particularly the 2015 release~\cite{undertaking2016european} and the 2017 Drone Outlook Study~\cite{sesar_2017}, the integration of UAVs into the European airspace gained momentum.

Regulations and operational restrictions have been introduced concurrently to ensure the safety of drone operations.
Notably, the EU regulation 2019/947~\cite{Ec2019} and Specified Operations Risk Assessment concepts~\cite{Easa2019} underscore the coexistence of unmanned and manned aircraft within shared airspace under rigorous safety checks.

To build an understanding of the research landscape concerned with such safe operations, we lay out publications linked to expressive and robust environment representations, planning approaches, and reasoning systems with the potential for jointly navigating these emerging societal rules and managing the physical demands of aerial mobility.

\subsection{Representation}
\label{sec:representation}

Robust and expressive environment representations are crucial for safe navigation.
From rural areas to urban canyons, they provide the necessary capabilities to manage knowledge about the agent's environment and co-inhabitants, as well as provide the necessary data to inform its routing (Section~\ref{sec:routing}) and reasoning (Section~\ref{sec:reasoning}) processes.

Numerous sensors can be employed by the UAV, specialized mapping platforms, or personal computing devices for crowd-sourcing data to gather the necessary environmental information.
They range from proprioceptive sensors, e.g., using the Global Navigation Satellite System~\cite{joubert2020developments} and Inertial Measurement Units~\cite{Zhang2012,Gao2022} which inform about the agent's own location, to exteroceptive sensors such as cameras~\cite{Alkendi2021,Flade2018}, Lidar~\cite{Demir2019,Petrlik2021}, and Radar~\cite{Ward2016,Michalczyk2022} which inform about the environment.

Frequently, maps created, e.g., from optical sensors~\cite{Levinson2010, Huang2019a, Lategahn2012} may lack deeper semantics and are unsuitable for navigation beyond simple obstacle avoidance, e.g., unstructured point clouds or unlabeled structures from image data.
To remedy this issue, vision-based classification approaches such as segment anything~\cite{Kirillov2023} and object-centric vision~\cite{locatello2020object,kipfconditional} can be employed to enrich the map with semantic labels for more principled navigation approaches.
Of course, human insights might also be employed to label the mapped environment features.

A prominent example of creating a semantically annotated, wide-coverage environment representation is the crowd-sourced project OpenStreetMap~\cite{Haklay2008}, storing its geospatial data in interconnected nodes alongside a rich tagging system.
Beyond static and quasi-static data, approaches such as the Relational Local Dynamic Map~\cite{Eggert2017} also represent the environment's transient and dynamic entities as a holistic, interconnected graph.

Recently, we have presented the Statistical Relational Map (StaR Map) as a method for encoding uncertain environments in a relational fashion~\cite{flade2024star}.
Rather than storing geometric information, StaR Maps consume data sources as discussed above to represent uncertain environments through sets of categorical and continuous parametric distributions, each associated with a spatial relation and a set of geographic features.
Hence, StaR Maps allow for explicit consideration of the inaccuracies of the mapping process and employed sensors with the annotated semantics of the respective environment features. 
Furthermore, they can be compiled into the parameters of probabilistic and logical reasoning frameworks such as our Constitutional Language (CoLa) as discussed in Section~\ref{sec:methods}.

\subsection{Routing}
\label{sec:routing}

Given sufficient knowledge about the agent's environment, decisions ought to be made on how to travel from the current position to a desired goal.
Although achieving the shortest time to fulfill a task is often the obvious target, many objectives may be imposed on the agent's path planning.

First, an agent may consider a single objective to decide a suitable path.
For example, risk minimization of collisions may be employed to decide velocity profiles that avoid getting too close to other traffic participants even under uncertainty~\cite{damerow2014predictive}.
Similarly, given more profound knowledge of the environment, such approaches can be enhanced by considering other agents' paths, allowing for more decisive and safe planning~\cite{puphal2023introducing}.

Since no singular optimal solution exists when considering multiple objectives simultaneously, a Pareto set of paths may be produced representing a range of trade-offs between the individual cost functions.
One early approach for many-objective path planning optimizes UAV paths for flights over rough terrain~\cite{nikolos2003evolutionary}. 
In a later approach, an urban setting in Singapore was adopted to plan 3D paths for UAVs~\cite{rubio2018data}. 
A special path representation was proposed, which was composed of the distance and the angle between waypoints and a straight line connecting the path's start and goal.

Evolutionary algorithms have been shown to successfully reduce the computational load of solving many-objective path planning~\cite{ghambari2020enhanced}.
To optimize a path regarding its length and risk, the NSGA-II algorithm~\cite{deb2002fast} was employed, a state-of-the-art evolutionary algorithm for multi-objective optimization capable of calculating a Pareto set.
Because using a grid-based path representation can result in sharp turns and inefficient jagged paths, a smooth spline curve representation is often preferred.
For a deeper dive into the topic, we refer the interested reader to an introduction on multi-objective optimization~\cite{emmerich2018tutorial}.

Our work is based on two core UAV routing ingredients: hybrid many-objective optimization for 3D path planning~\cite{hohmann2023three} and the Probabilistic Mission Design (ProMis) framework~\cite{kohaut2023mission}.
While the former builds on a graph-based search and a subsequent evolutionary search for fine-granular routing, the latter shows how reasoning on legal frameworks and operational preferences can be enabled via hybrid probabilistic logic programs, producing independent and identically distributed probabilities of the requested model being satisfied at each continuous point in state space.

By combining these two frameworks, a powerful UAV routing pipeline emerges that consumes a hybrid probabilistic, logical representation of legal constraints alongside physical objectives to produce a Pareto set of proposal paths.
This not only yields compliant and effective paths but also allows for individual trade-offs between objectives without the need to enforce each with the same strictness.
Furthermore, we demonstrate how mission design tasks such as Clearance, Explanation, and Optimization of proposal path and mission parameters are supported~\cite{kohaut2024ceo}, forming an interpretable, adaptable, and safe navigation system.

\subsection{Reasoning}
\label{sec:reasoning}

When creating autonomous mobile agents, it is necessary to consider their ability to reach their goal without collisions and simultaneously reason on the circumstances to conclude safe and compliant decisions.
Endowing an agent with such capabilities, e.g., ensuring that local legal regulations are satisfied, is subject to ongoing research.

Recent research endeavors have enhanced safety analysis and risk assessment methodologies. 
For instance, Rothwell and Patzek~\cite{RothwellPatzek2019} have contributed by employing satisfiability checks on symbolic models to verify and improve mission planning for UAVs. 
In contrast, Rakotonarivo et al.~\cite{Rakotonarivo2022} highlight the importance of directly fitting the output of safety analysis and risk assessment to map or environmental data. 
More specifically, they propose to improve the visual representation and enable data exploration by displaying interactive representations of mission parameters. 
This approach improves integration with environmental data, positively impacting the effectiveness of safety analysis and risk assessment in real-world scenarios.

Several studies, such as~\cite{Primatesta2020} and~\cite{Raballand2021}, have introduced approaches that integrate visual maps with risk models to calculate and visually represent risks associated with UAV operations. 
These models prioritize risk assessment based on formal frameworks, identifying potential hazards and safety concerns, especially ground casualties and transportation network disruptions.

In this work, we build on ProMis and its capabilities for facilitating crucial tasks such as mission Clearance, Explanation, and Optimization~\cite{kohaut2024ceo}.
It aligns with preceding research efforts, offering a formal symbolic approach to probabilistic classification and mission design through spatial reasoning based on systems such as StaR Maps~\cite{flade2024star}. 
Notably, ProMis provides an adaptable and interpretable interface to mission design while allowing arbitrary queries such as those about the satisfaction of local airspace regulations, thus facilitating comprehensive compliance assessment and decision-making.
Moreover, ProMis accommodates evolving regulatory demands and operational priorities within its probabilistic framework, such as battery lifetime~\cite{An2023} or weather effects~\cite {Schuchardt2022}.
By extending ProMis with CoLa as a high-level modeling language, we move beyond its original focus on compliance and introduce physical objectives into the model, allowing ProMis' application for real-world UAV routing with path effectiveness in mind.

    \section{Methods}
\label{sec:methods}

\subsection{Architecture}

\begin{figure}[t]
    \centering
    \includegraphics[width=\linewidth]{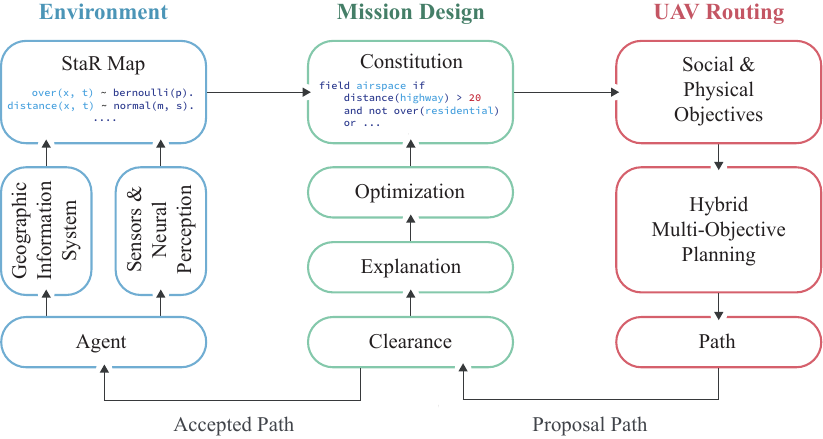}
    \caption{
        \textbf{An architecture for effective and compliant UAV routing through Probabilistic Mission Design and Hybrid Many-Objective Optimization:}
        Our architecture is made up of three pillars.
        First, a hybrid probabilistic and relational environment representation based on StaR Maps.
        Second, a probabilistic mission design framework based on ProMis, where rules of the shared traffic space and physical objectives to be considered are encoded in the agent's Constitution to feed into the routing and allow for Clearance, Explanation, and Optimization of a proposed path.
        Third, an evolutionary many-objective UAV routing system producing Pareto sets of proposal paths.
    }
    \label{fig:architecture}
\end{figure}

We begin by outlining our architecture for compliant and effective Unmanned Aerial Vehicles (UAV) routing.
It can be divided into the three pillars of environment representation, mission design, and UAV routing through many-objective optimization.
We base each of these pillars on our prior publications on Statistical Relational Maps (StaR Maps)~\cite{flade2024star}, Probabilistic Mission Design (ProMis) with Clearance, Explanation and Optimization techniques~\cite{kohaut2023mission, kohaut2024ceo}, and Hybrid Many-Objective Optimization for 3D UAV routing~\cite{hohmann2023three} respectively.
The resulting architecture for such effective and compliant UAV routing is illustrated in Figure~\ref{fig:architecture}.

We assume a UAV with access to data from a Geographic Information System, such as the local road network, buildings, and infrastructure, as well as (neural) sensors for online perception of more dynamic environment features such as people or other vehicles.
With this information at hand, the agent can compute and maintain a StaR Map (see Section~\ref{sec:modeling_regulations}) to represent its environment in a hybrid probabilistic set of spatial relations between the agent's state space and the environment, e.g., probabilities wether the agent would be \textit{over} a certain feature or at a required \textit{distance} in a specific state.

The StaR Map is employed to parameterize the agent's Constitution (see Section~\ref{sec:cola}), i.e., a unified description of the applicable local airspace regulations and physical objectives to be considered during travel.
While physical objectives are provided in analytical models (see Section~\ref{sec:physical_objectives}), inference in probabilistic first-order logic allows the agent to estimate how likely states across the navigation space will satisfy regulations.

Based on these objectives, we employ the Versatile Intelligent Aerial Streets (VIAS) routing algorithm, resulting in a Pareto front of paths (see Section~\ref{sec:vias}).
Given a weighting of the objectives, a proposed path is decided and checked for Clearance based on the traffic restrictions encoded in the Constitution (see Section~\ref{sec:clearance}).
If Clearance is granted, the agent accepts and executes the path.
Otherwise, an Explanation can be given to understand which mission parameters impacted the decision on Clearance the most (see Section~\ref{sec:explanation}), and the Constitution's parameters may be searched for a more suitable setting, e.g., changing at what time of day the mission shall be carried out (see Section~\ref{sec:optimization}).

\subsection{Statistical Relational Maps for Reasoning on Airspace Regulations}
\label{sec:modeling_regulations}

We have recently introduced StaR Maps to encapsulate uncertain environments of semantically annotated features, i.e., geometry with associated stochastic error parameters and labeled with descriptive tags~\cite{flade2024star}.
Rather than containing a graphical representation of the environment, a StaR Map parameterizes hybrid probabilistic relations between points in state space and sets of features.

Consider the following two models of spatial relations: 
\begin{align}
    f:&\ \mathbb{R}^d \times \mathbb{T} \times \mathbb{M} \rightarrow \mathbb{N} \\
    g:&\ \mathbb{R}^d \times \mathbb{T} \times \mathbb{M} \rightarrow \mathbb{R}
\end{align}
According to StaR Maps, a spatial relation with domain and codomain of $f$ describes a categorical relationship between a point in $\mathbb{R}^d, d \in \mathbb{N}$, a tag in $\mathbb{T}$ referencing a set of map features, and a map in $\mathbb{M}$.

For example, such a relation may categorize points into whether they lie \textit{over} any feature with the respective tag.
Analogously, a relation with domain and codomain of $g$ describes a quantitative relationship, e.g., the \textit{distance} between a point and the closest element in the set of features with the respective tag.
In turn, the set $\mathbb{T}$ itself depends on the application, e.g., in the UAV routing context, one might have $\mathcal{T} = \{park, road, pilot, building\}$ to reference important environment features that an agent may avoid flying over or keep a certain distance to.

Let a Map $\mathcal{M} = (\mathcal{V}, \mathcal{E}, t) \in \mathbb{M}$ be a triple of vertices $\mathcal{V}$, edges $\mathcal{E}$ and tagging function $t$.
If a path exists between two vertices in $\mathcal{V}$ across edges in $\mathcal{E}$, we consider them part of the same feature.
For each vertex $\mathbf{v} \in \mathcal{V}$, the function $t(\mathbf{v}) \subseteq \mathbb{T}$ annotates $\mathbf{v}$ with a set of semantic tags.

StaR Maps introduced Uncertainty Annotated Maps (UAM) $\mathcal{U} = (\mathcal{M}, a, b)$ as a triple of a map $\mathcal{M}$ and two annotator functions $a$ and $b$ that assign transformation parameters $\vec{\alpha} = a(\vec{v})$ and translation parameters $\vec{\beta} = b(\vec{v})$ of a stochastic error model~\cite{flade2021error}.
Based on a UAM, we sample $n_M \in \mathbb{N}$ times the parameters of an affine map such that 
\begin{align}
    \Phi^{(n_M)} &\sim \phi(a(\vec{v})) \text{ and } \\
    \vec{o}^{(n_M)} &\sim \kappa(b(\vec{v}))
\end{align}
can be employed to derive from an original vertex $\vec{v}$ the sample
\begin{align}
    \vec{v}^{(n_M)} &= \Phi^{(n_M)} \cdot \vec{v} + \vec{o}^{(n_M)}.
\end{align}
Here, $\phi$ is a distribution over matrices $\Phi^{n_M}$ to apply geometric transformations, e.g., rotations, that keep the center point fixed. 
At the same time, $\kappa$ generates offset vectors $\vec{o}_m$ to apply a translation.
From this process, one obtains a collection $\mathcal{W} = \{\mathcal{M}^{(0)}, ..., \mathcal{M}^{(n_M)}\}$ of perturbed maps.

As the evaluation of a deterministic spatial relation $r$ will differ for each perturbed map, StaR Maps computes the statistical moments to fit the distributions chosen for each relation. 
To estimate the parameters for a specific spatial relation $r$ and a type $t$ on a point $\vec{x} \in \mathbb{R}^d$, we compute $\rho^{(n_M)} = r(\vec{x}, t, \mathcal{M}^{(n_M)})$.
With the set $\mathcal{P}_r = \{\rho^{(0)}, ..., \rho^{(n_M)}\}$ at hand, we obtain the statistical moments, e.g., mean and variance, of $\mathcal{P}_r$.

Using maximum likelihood estimation or moment matching with the desired distribution, for example, a Bernoulli distribution for $f$- or a Gaussian distribution for $g$-domain relations, we store the resulting parameters within a StaR Map.
The so-obtained spatial relations can then be compiled into distributional ground atoms in probabilistic first-order logic, which is how we proceed in Section~\ref{sec:cola}.

\subsection{Physical Objectives}
\label{sec:physical_objectives}

In practice, agents ought not only to obey traffic laws but also to carefully consider and pursue various physical objectives as well.
Here, we consider exemplary physical and geo-referenced objective function formulations evaluating a path regarding the risk of injury for city residents, noise emission, radio signal disturbance, and energy consumption, which have been presented in detail in a prior publication~\cite{hohmann2023three}.
Hence, in the following, we re-introduce these objective functions briefly and refer to the original work for an in-depth discussion. 

We categorize and discuss these objectives in two forms: grid-based and non-grid-based.

Grid-based objectives are provided as three-dimensional discrete scalar functions of the form
\begin{equation}
	\mathcal{G}^{\mathrm{3D}}:\underbrace{\{1,\ldots,\left\lfloor|\Delta x|/x_\mathrm{res}\right\rfloor\}\times\{1,\ldots,\left\lfloor|\Delta y|/y_\mathrm{res}\right\rfloor\}\times\{1,\ldots,\left\lfloor|\Delta z|/z_\mathrm{res}\right\rfloor\}}_{\mathbb{D}^{\mathrm{3D}}}\rightarrow \mathbb{R}^+,
\end{equation} 
with $\Delta x = x_\mathrm{max}-x_\mathrm{min}$, $\Delta y = y_\mathrm{max}-y_\mathrm{min}$, and $\Delta z = z_\mathrm{max}-z_\mathrm{min}$. 
Hence, a grid-based objective function is defined through the discrete line integral 
\begin{equation}
	f_\mathcal{G}(\Pi) = \sum_{i=1}^{|\Pi|-1}\frac{\mathcal{I}(\mathcal{G}^{\mathrm{3D}},\pi_{i-1})+\mathcal{I}(\mathcal{G}^{\mathrm{3D}},\pi_i)}{2}|\pi_i - \pi_{i-1}|,\label{eq:line_integral}
\end{equation} 
over the scalar function along the sequence of waypoints (i.e., path) $\Pi = [\pi_0, \ldots, \pi_{|\Pi|-1}]$, where $\mathcal{I}$ is an interpolation function.
We have presented different such scalar functions $\mathcal{G}^{\mathrm{3D}}$ (i.e., grid maps) to yield distinct objective functions.

First, the \textit{risk} grid map assigns lower risk values to grid cells over buildings and water areas, assuming a reduced likelihood of human injury in case of a UAV malfunction. 
Risk values increase with altitude.

Second, the \textit{noise} grid map assigns lower noise values to grid cells over city roads, as drone-generated noise blends with traffic noise and is less likely to disturb city residents. 
Noise values decrease with altitude.

Third, the \textit{radio disturbance} grid map assigns lower disturbance values to grid cells near radio cell towers. 
Disturbance values increase with distance from these towers, following the inverse square law 
\begin{equation}
    D(\vec{p}) = \frac{D_0}{(\mu r + 1)^2},
    \label{eq:radio}
\end{equation} 
with  $r = |\vec{p}_{\mathrm{R}}-\vec{p}|$ being the Euclidean distance between the grid cell position $\vec{p}\in\mathbb{D}^{\mathrm{3D}}$ and the radio cell tower position $\vec{p}_{\mathrm{R}}\in\mathbb{D}^{\mathrm{3D}}$, $D_0\in\mathbb{R}^-$ being the best signal value, and $\mu\in\mathbb{R}^+$ being a scaling factor.

We refer to cost functions that cannot be modeled using Equation~\ref{eq:line_integral} as non-grid-based cost functions.
This includes our model for \textit{energy consumption}
\begin{equation}
    f_\mathrm{E}(\Pi) = \frac{1}{2}m v_{xy}^2 + c_\mathrm{E}\left(|\Pi_{xy}| + 10|\Pi_{z,\uparrow}| + 15|\Pi_{z,\downarrow}|\right),
    \label{eq:energy_model}
\end{equation}
where $|\cdot|$ denotes the Euclidean length of the path projections in the $xy$-plane $\Pi_{xy}$, the upward-pointing $z$-direction $\Pi_{z,\uparrow}$, and the downward-pointing $z$-direction $\Pi_{z,\downarrow}$, $m$ is the UAV mass, $v_{xy}$ is the cruise velocity and $c_\mathrm{E}$ denotes a vehicle-specific energy coefficient.

\subsection{The Constitutional Language}
\label{sec:cola}

In this work, we introduce the Constitutional Language (CoLa) for UAV routing, i.e., a probabilistic and symbolic, unified description language encoding all of its navigational constraints and objectives in one place.

A Constitution written in CoLa, as exemplified in Listing~\ref{listing:constitution} is written in three parts:
First, one defines the setting by referring to a StaR Map, providing the probabilistic spatial relations one may use to encode the legal requirements, and determining the parameters under which the mission may be carried out, e.g., weather conditions, licensing information, or abstract UAV properties.
Second, one defines further background knowledge and rules on which the compliance of an agent's state can be reasoned.
Third, one defines which quantities shall be treated as objectives across the navigation space.
To this end, CoLa refers to \textit{field} clauses and objectives if they relate to functions that produce scalar fields across the agent's state space.

While on the one hand, physical objectives as discussed in Section~\ref{sec:physical_objectives} may be incorporated through their mathematical description in a language such as Python~\footnote{https://www.python.org/}, the logical reasoning on airspace regulations is facilitated by compiling the relevant parts of the Constitution into a probabilistic logic programming language supporting hybrid probabilistic inference, i.e., on discrete and continuous distributions simultaneously, such as Hybrid ProbLog~\cite{nittihybrid}.
Other systems with more advanced neuro-symbolic capabilities, such as DeepProbLog~\cite{deepproblog}, NeurASP~\cite{neurasp}, or SLASH~\cite{slash}, are left for future work, e.g., to integrate end-to-end learning with deep learning models.

\begin{listing}
    \caption{
        \textbf{Programming an agent's Constitution:}
        Here, we show an example of the Constitution, where a simple model for local airspace regulations and a set of physical objectives are imposed on the UAV's path planning.
        Note that the choice of parameters is not made within the Constitution but at runtime through the operator or automatically selected by maximizing a mission's probability of satisfying all requirements. 
    }%
    \label{listing:constitution}%
    \centering
    \begin{minted}
    [
        frame=none,
        autogobble,
        fontsize=\footnotesize,
        xleftmargin=20pt,
        linenos
    ]{python}
    # Mission setting and parameter options
    star_map("./environments/example.star").      # StaR Map according to Section 3.2
    parameter {regular_license, special_license}. # Up to choice of operator or mission
    parameter {day, night}.                       #    optimization search (Section 3.8)

    # Airspace rules using hybrid probabilistic spatial relations from StaR Map 
    take_off_mass ~ normal(20.0, 1.0).
    light_drone if take_off_mass < 5.0.
    field line_of_sight if day and distance(pilot) < 100.
    field airspace if line_of_sight and regular_license and take_off_mass < 10.0
        or special_license and take_off_mass < 25.0.

    # The UAV's objectives
    field objective airspace.                     # Defined by probabilistic logic
    field objective radio("./models/radio.py").   # Defined by Python implementations
    path objective energy("./models/energy.py").  #    of Section 3.3
    \end{minted}
\end{listing}

Each Constitution contains a first-order logic theory annotated with categorical and continuous probability distributions.
To this end, a Constitution draws parameters from a StaR Map, e.g., categorical distribution $over(\cdot)$ and continuous distribution $distance(\cdot)$ to allow probabilistic spatial reasoning, which is a core requirement of representing many relevant airspace regulations.
For example, requirements may be that a UAV shall not fly $over(bystander)$ and with $distance(operator) < 20$.

To query for the probability that the Constitution's modeled requirements are satisfied for a location $\vec{x} \in \mathbb{R}^d$, it needs to be compiled into the chosen backend, such as Hybrid ProbLog~\cite{nittihybrid}.
This is achieved by (i) replacing syntactic elements, e.g., conjunctions (\textit{and}) are written as comma-separated lists in ProbLog, and (ii) inserting the parameters of the StaR Map.
We refer to the original paper for details on the latter part~\cite{flade2024star}.

Such a probabilistic logic program must first be solved, i.e., the models under which it is satisfied must be found to perform probabilistic inference.
While details will differ across off-the-shelf solving pipelines, the rough process can be described in two steps.
First, the program is grounded, meaning all variables are substituted, and the program is rewritten in Conjunctive Normal Form (CNF).
Second, the actual solver consumes the CNF, producing a Disjunctive Normal Form (DNF) where every conjunction represents a possible assignment $j \in \mathcal{J}$ of ground atoms $a \in \mathcal{A}$ under which the (constrained for the currently queried objective) program is satisfied.
For instance, one may employ clingo~\cite{gebser2017clingo}, a state-of-the-art tool for solving programmatic first-order logic based on Answer Set Programming~\cite{brewka2011answer}.

For exact probabilistic inference, one needs to assign the probabilities $P(a = j(a) | \vec{x}_t)$ of atom $a \in \mathcal{A}$ to take on the value assigned by model $j \in \mathcal{J}$.
These parameters have been either written down directly in the Constitution, e.g., the take-off mass in Listing~\ref{listing:constitution}, or are provided by the StaR Map.
In turn, the probability $P(SAT | \vec{x})$ of the Constitution being satisfied given the state and measurement at time $t$ is then obtained via the sum-product
\begin{align}
    P(SAT | \vec{x}) = \sum\nolimits_{j \in \mathcal{J}} \prod\nolimits_{a \in \mathcal{A}} P(a = j(a) | \vec{x}).
    \label{eq:wmc}
\end{align}
A knowledge compiler is often employed to compress this sum-product using a heuristic search for a minimal formula, often leading to substantial speedups of the inference~\cite{muise2012d}.

Note that the parameters and the set of points for which the Constitution is evaluated are a choice made when evaluating the model for routing and are not directly encoded within the Constitution itself.
If multiple compliance objectives are stated in the provided CoLa program, their conjunction is queried instead and passed on to the path planner.

\subsection{Many-Objective Path Planning}
\label{sec:vias}

We facilitate the search for a Pareto set of paths considering the objectives stated in a CoLa program through, e.g., the physical objectives defined in Section~\ref{sec:physical_objectives} and airspace-dependent regulations stated in probabilistic first-order logic by employing the Versatile Intelligent Aerial Streets (VIAS) framework for many-objective path planning as presented in a prior publication~\cite{hohmann2021hybrid}.
Besides the reference implementation of the overall architecture (see Figure~\ref{fig:architecture}) being made available as an open-source package, VIAS is also available as a stand-alone solution.

In its first stage, VIAS applies the single-objective Dijkstra algorithm to $n_\mathrm{S}$ weighted aggregations of all grid-based objective functions. 
This results in $n_\mathrm{S}$ three-dimensional polygonal paths guaranteed optimal in the employed grid structure.
Next, VIAS smooths and approximates the polygonal paths and obtains three-dimensional NURBS curves~\cite{nurbs_book_nurbs, nurbs_book_approximation}.
These splines are henceforth treated as initial solutions in the second stage, a many-objective evolutionary optimization problem calculating an $E$-dimensional Pareto set of smooth paths that can be thought of as a set of optimal trade-off solutions between all considered objective functions.

Evolutionary algorithms are particularly suitable for finding non-dominated solutions and approximating a diversified Pareto front without needing a weighted aggregation of the objective functions. 
Besides, the Dijkstra algorithm provides optimal solutions on a discrete, i.e., graph-based path representation and is computationally efficient (compared to a metaheuristic search). 
It has been shown to yield suitable initial solutions to evolutionary algorithms despite the differences in path representation~\cite{hohmann2022multi}. 
Thus, the hybrid combination of both approaches combines their strengths, speeding up the evolutionary search in VIAS.

The smoothing and approximation between the two optimization stages are crucial.
At this point, the path representation changes from a polygonal path definition to a NURBS definition.
Here, we will only provide a brief discussion; the interested reader is referred to~\cite{hohmann2023three}.

VIAS first smoothes the polygonal path by applying a Gaussian filter with kernel $\mathcal{K} = \begin{bmatrix} 1 & 2 & 1 \end{bmatrix}$.
Then, the smoothed paths are approximated in the form of a NURBS curve.
For this operation, the critical parameter is the number of control points, $n_\mathrm{P}$, for the resulting NURBS curve. 
While a low number $n_\mathrm{P}$ may lead to an excessive approximation error, a large number $n_\mathrm{P}$ increases the size of the optimization vector, expanding the search space during the evolutionary optimization step.

Hence, the optimal number of control points depends on the length and curvature of the Dijkstra path, making it impractical to pre-define $n_\mathrm{P}$ as a hyperparameter. Instead, we adaptively determine the appropriate number of control points for each scenario. 
This adaptive process starts with a minimum value and successively increments the number of control points. 
Each iteration involves:
\begin{enumerate}
    \item Approximating the NURBS curve with the current number $n_\mathrm{P}$.  
    \item Calculating the error between the Dijkstra path and its NURBS approximation.  
\end{enumerate} 
The process concludes when the error falls below a specified threshold, indicating a suitable number of control points for accurate approximation, yielding an efficient optimization of Pareto-optimal paths to be considered for the UAV.

\subsection{Mission Clearance}
\label{sec:clearance}

Given a CoLa program and the subsequently obtained Pareto set of paths from VIAS as outlined in Sections~\ref{sec:cola} and \ref{sec:vias} respectively, one may still wonder if the results are overall sufficiently likely to satisfy the modeled airspace regulations.
Hence, based on our prior publication on Clearance, Explanation, and Optimization in ProMis~\cite{kohaut2024ceo}, we discuss each of these steps with regard to Pareto-optimal proposal paths.

Assume a proposal path $\Pi = [\pi_0, \ldots, \pi_{|\Pi|-1}]$ was selected to be considered for the UAV's next journey.
To decide clearance, we pass each waypoint into the CoLa program to query the respective probabilities of satisfying the model.
Then, clearance is granted if the average probability
\begin{equation}
    C = \frac{1}{|\Pi|} \sum_i P(\mathrm{SAT} | \pi_i)
    \label{eq:clearance}
\end{equation}
is greater than a set threshold $T_C$.

Note that this Clearance does not necessarily coincide with a path being the knee-point or even optimal with regard to the airspace requirements.
On the other hand, an extreme point of the Pareto set for an unrelated objective may still be granted Clearance, i.e., this step has to be performed on top of the many-objective routing itself.

If Clearance is denied, the next step may be to Explain which parameters of the Constitution are at fault and may be altered to Optimize the mission setting.

\subsection{Mission Explanation}
\label{sec:explanation}

Explaining (probabilistic) models aims to understand how inputs correspond to outputs.
A model explainer approximates the model's behavior, highlighting how each input pushes the output towards a label.
In the case of Clearance for UAV routing, we can gain insights into which parameters of the Constitution lead to a route being rejected or accepted.
In other words, we aim not just to produce a definitive answer to a query but to understand why the answer was generated and how to best sway the setting for Clearance.

Consider again the example CoLa program in Listing~\ref{listing:constitution}.
Here, two sets of parameters were defined, one on the employed license and one on the time of day.
With regard to granting clearance, it is easy to see in this case that a special operations license allows for easier Clearance since no requirement on visual line-of-sight is imposed.

In the Explanation step, the space of combinations of parameters is probed, each producing an individual score for the given path.
With this variation, the operator can be informed on what changes to the parameters of their mission may be suitable and what their impact on the prior decision about Clearance is.

\subsection{Mission Optimization}
\label{sec:optimization}

While Clearance informs about the acceptance of a proposed path and mission parameters, and Explanation gives insight into the impact of alternative mission parameters, searching for the optimal setting allows the exploitation of free parameter choices to find the most compliant setting.

Given the results of the Explanation stage, the optimal setting is chosen such that the Clearance score of Equation~\ref{eq:clearance} is maximized for the proposed path.
Note that this may allow running journeys that were initially unsuitable due to local airspace restrictions.
Hence, solutions that are less careful about satisfying the restrictions may prove appropriate for Clearance, e.g., effective paths that are better at conserving the energy of the UAV.

\begin{table}
    \caption{Parameters for environment representation, reasoning, objectives, and  routing.}
    \begin{tabular}{cllr}
        \toprule
        & Name & Symbol & Value \\ 
        \midrule 
        \multirow{5}{*}{
            \shortstack{Navigation\\Space}} & Origin & $\lambda, \phi$ & $48.8677\text{°N}, 2.3391\text{°E}$ \\
            & Easting Range & $\left[x_\text{min}, x_\text{max}\right]$ & $\left[\SI{0}{\meter}, \SI{13000}{\meter}\right]$ \\
            & Northing Range & $\left[y_\text{min}, y_\text{max}\right]$ & $\left[\SI{0}{\meter}, \SI{13000}{\meter}\right]$ \\
            & Altitude Range & $\left[z_\text{min}, z_\text{max}\right]$ & $\left[\SI{0}{\meter}, \SI{300}{\meter}\right]$ \\
            & Discretization & $x_\text{res}, y_\text{res}, z_\text{res}$ & $\SI{10}{\meter}, \SI{10}{\meter}, \SI{10}{\meter}$ \\
        \midrule
        \multirow{4}{*}{StaR Map}
            & Data source && OpenStreetMap~\cite{OpenStreetMap}  \\
            & Translation error & $\vec{t}$ & $\mathcal{N}(\SI{0}{\meter}, \SI{9}{\meter\squared})$ \\
            & Number of sampled maps & $n_\mathrm{M}$ & $50$ \\
            & Interpolator && Linear \\
        \midrule
        \multirow{2}{*}{ProMis}
            & Legal Model & $\mathcal{C}$ & Listing~\ref{listing:paris_constitution} \\
            & Clearance threshold & $T_C$ & Variable \\ 
        \midrule 
        \multirow{3}{*}{UAV} 
            & Mass & $m$ & $\SI{1.2}{\kilo\gram}$ \\ 
            & Cruise velocity & $v_{xy}$ & $\SI{14}{\meter\per\second}$\\
            & Energy coefficient & $c_\mathrm{E}$ & $\SI{9.12}{\joule\per\meter}$ \\
        \midrule
        \multirow{3}{*}{Radio} 
            & Cell height & $z_\mathrm{R}$ & $\SI{75}{\meter}$ \\ 
            & Initial radio signal disturbance & $D_0$ & -200 \\
            & Scaling factor & $\mu$ & $1.0$ \\
        \midrule
        \multirow{4}{*}{Path} 
            & Num. control points & $n_\mathrm{P}$ & Variable \\
			& Basis function degree & $d$ & $2$ \\
			& Parametrization & & Chordal \\
			& Waypoint resolution & $\Delta \pi$ & $\SI{5}{\meter}$ \\
        \midrule
        \multirow{9}{*}{\shortstack{VIAS\\Optimizer}}
            & Optimization vector dimension & $D$ & Variable \\
			& Mutation operator && Gaussian \\
            & Mutation stepsize & $\vec{\sigma}$ &  $\left[\SI{10}{\meter}, \ldots, \SI{10}{\meter}\right] \in\mathbb{R}^D$\\
            & Mutation probability & $P(M)$ & $1.0$ \\
            & Individual mutation probability & $P(M_i)$ & $1/D$ \\
            & Crossover operator &  & One Point \\
            & Crossover probability & $P(C)$ & $0.9$ \\
			& Number of individuals & $n_\mathrm{I}$ & $700$ \\
            & Number of weighted solutions & $n_\mathrm{S}$ & $70$ \\
        \bottomrule
    \end{tabular}
    \label{tab:parameters}
\end{table}

Of course, the optimal choice in parameters may not be an option under the respective circumstances, e.g., no pilot with a more advanced license may be available at that time.
Hence, the search space may be further restricted at runtime.

    \section{Experiments}
\label{sec:experiments}

Throughout Section~\ref{sec:methods}, we have presented an approach for effective and compliant UAV routing under many objectives.
In this Section, we evaluate our methods on real-world, crowd-sourced data from the city of Paris, demonstrating how our framework not only jointly models airspace regulations and physical objectives but also produces suitable Pareto sets trading-off the objectives and applying Clearance, Explanation, and Optimization on the mission setting to further improve on the found solutions.

\begin{listing}
    \caption{
        \textbf{A Constitution for navigating the skies over Paris:}
        We employ a StaR Map sourced from crowd-sourced OpenStreetMap data to describe a simplified set of airspace regulations applied to the urban environment.
        Here, while government offices and embassies always require a safety distance for UAVs, the restrictions with regard to the other considered types of map features are altitude, license, and time-dependent. 
        With regard to physical objectives, we consider radio disturbance, noise pollution, risk, and energy consumption alongside legal compliance.
    }%
    \label{listing:paris_constitution}%
    \centering
    \begin{minted}
    [
        frame=none,
        autogobble,
        fontsize=\footnotesize,
        xleftmargin=20pt,
        linenos
    ]{python}
    # Which StaR Map to use
    star_map("./environments/paris.star").
    
    # Mission parameters describing the circumstances
    parameter {standard_license, expanded_license}.
    parameter {daytime, nighttime}.
    
    # Limitationss depending on altitude
    field low_flight_limitations if over(park) 
        or distance(primary) < 30 
        or distance(secondary) < 15.
    field mid_flight_limitationss if low_flight_limitations 
        or distance(building) < 20.
    field high_flight_limitations if mid_flight_limitations 
        or daytime and distance(stadium) > 50 and distance(stadium) < 150.
    
    # Altitude independent government limitations
    field government_limitations if distance(government) > 200 and distance(embassy) > 200.
    
    # The satisfying airspace over Paris
    field objective paris_limitations if expanded_license and government_limitations
        or standard_license and government_limitations and (
               altitude < 100 and low_flight_limitations
            or altitude < 200 and medium_flight_limitations
            or altitude < 300 and high_flight_limitations
        ).
    
    # Physical objectives for effective routing
    field objective radio("./models/radio.py").
    field objective noise("./models/noise.py").
    field objective risk("./models/risk.py").
    path objective energy("./models/energy.py").
    \end{minted}
\end{listing}

\subsection{Scenario Definition}

For our experiments, we consider a large-scale urban environment with data sourced from OpenStreetMap~\cite{OpenStreetMap} and OpenCellID~\cite{ocid} for modeling airspace regulations and physical objectives.
As crowd-sourced data is untrustworthy, we assume a uniform translational error on all map features.
Table~\ref{tab:parameters} lists the parameters we use throughout our experiments.
We draw data from a $13 \times \SI{13}{\kilo\meter\squared}$ area, covering Île-de-France, with legal and physical requirements stated in the CoLa program in Listing~\ref{listing:paris_constitution}.
Furthermore, we split the area into $\SI{1}{\kilo\meter\squared}$ tiles where, in each one, two paths from the respective opposing corners ought to be found, starting and ending at ground level, forming a large-scale network across the metropolitan area.

\subsection{A Constitution for Urban Advanced Aerial Mobility}

\begin{figure}
    \begin{subfigure}[b]{0.23\linewidth}
        \raisebox{5.75mm}{\includegraphics[width=\linewidth]{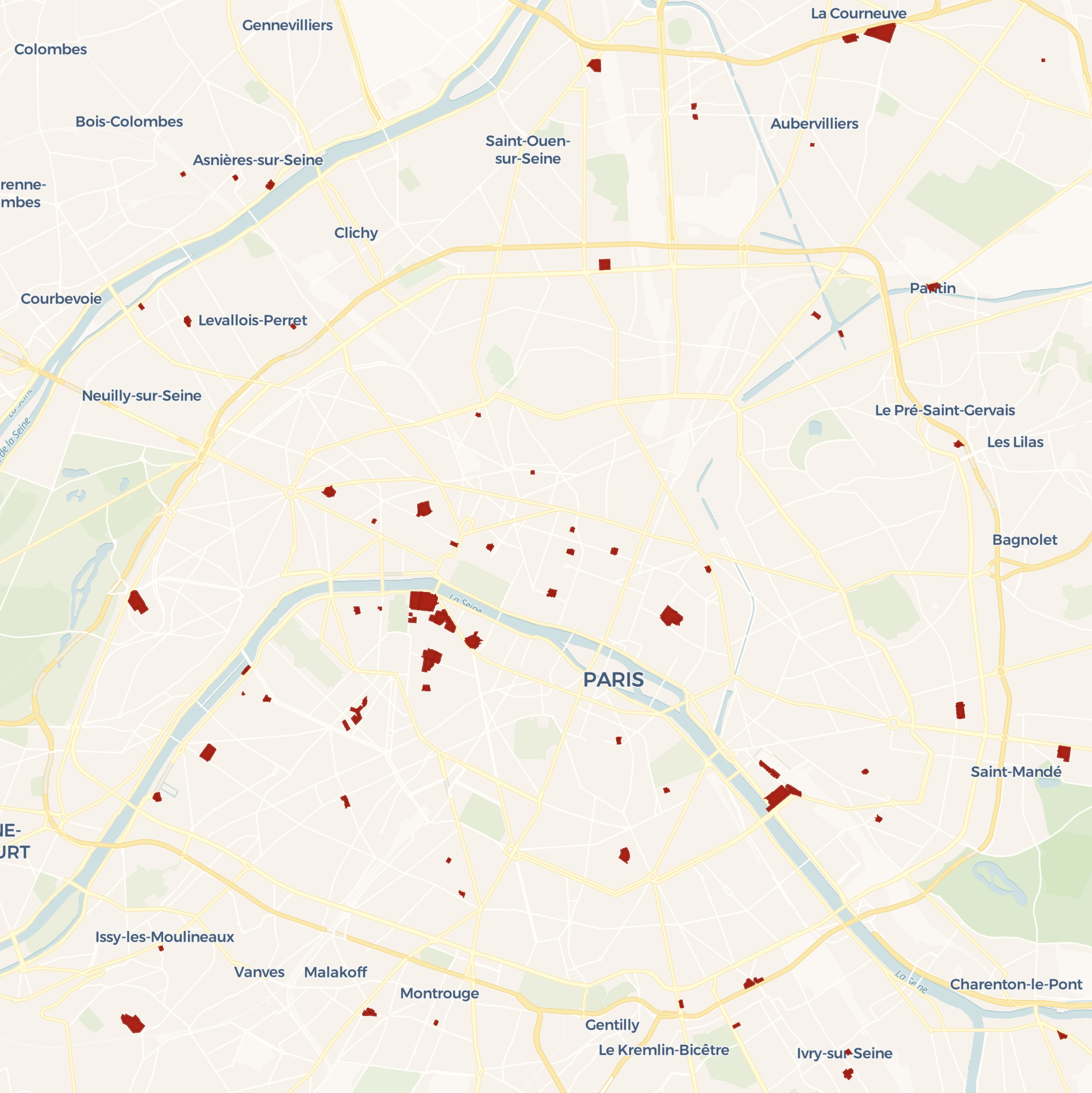}}
        \caption{Government buildings}
    \end{subfigure}
    \hfill
    \begin{subfigure}{0.35\linewidth}
        \includegraphics[width=\linewidth]{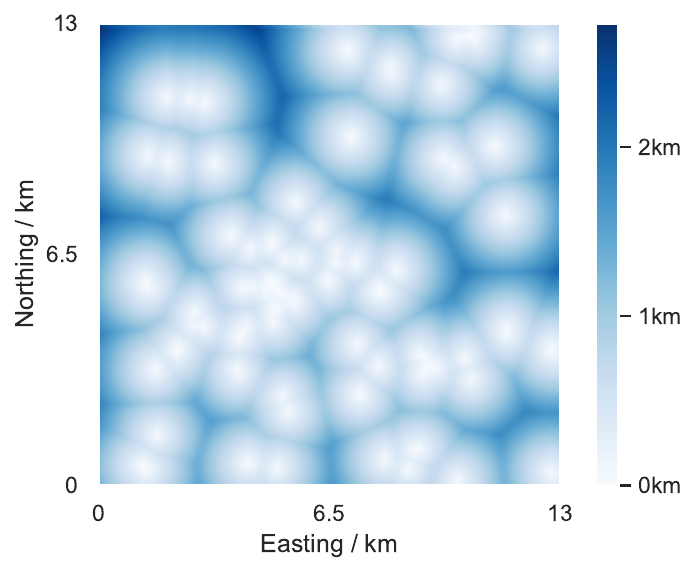}
        \caption{$\mu$ of \textit{distance(government)}}
    \end{subfigure}
    \hfill
    \begin{subfigure}{0.35\linewidth}
        \includegraphics[width=\linewidth]{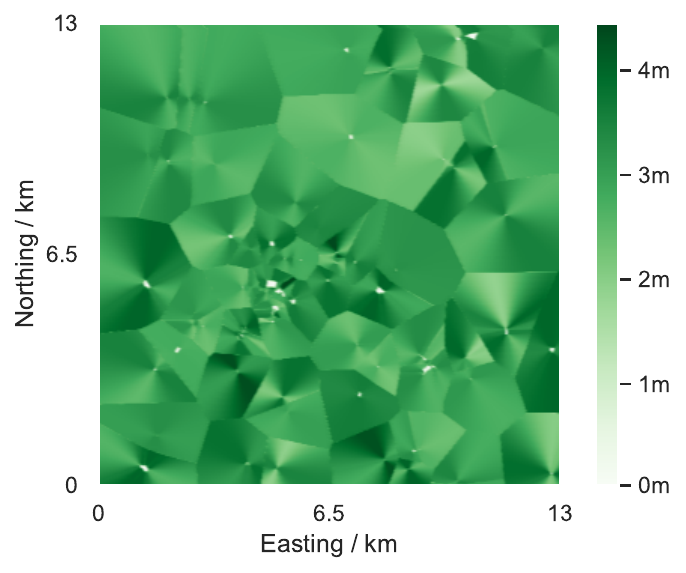}
        \caption{$\sigma$ of \textit{distance(government)}}
    \end{subfigure}\\
    \begin{subfigure}[b]{0.23\linewidth}
        \raisebox{5.75mm}{\includegraphics[width=\linewidth]{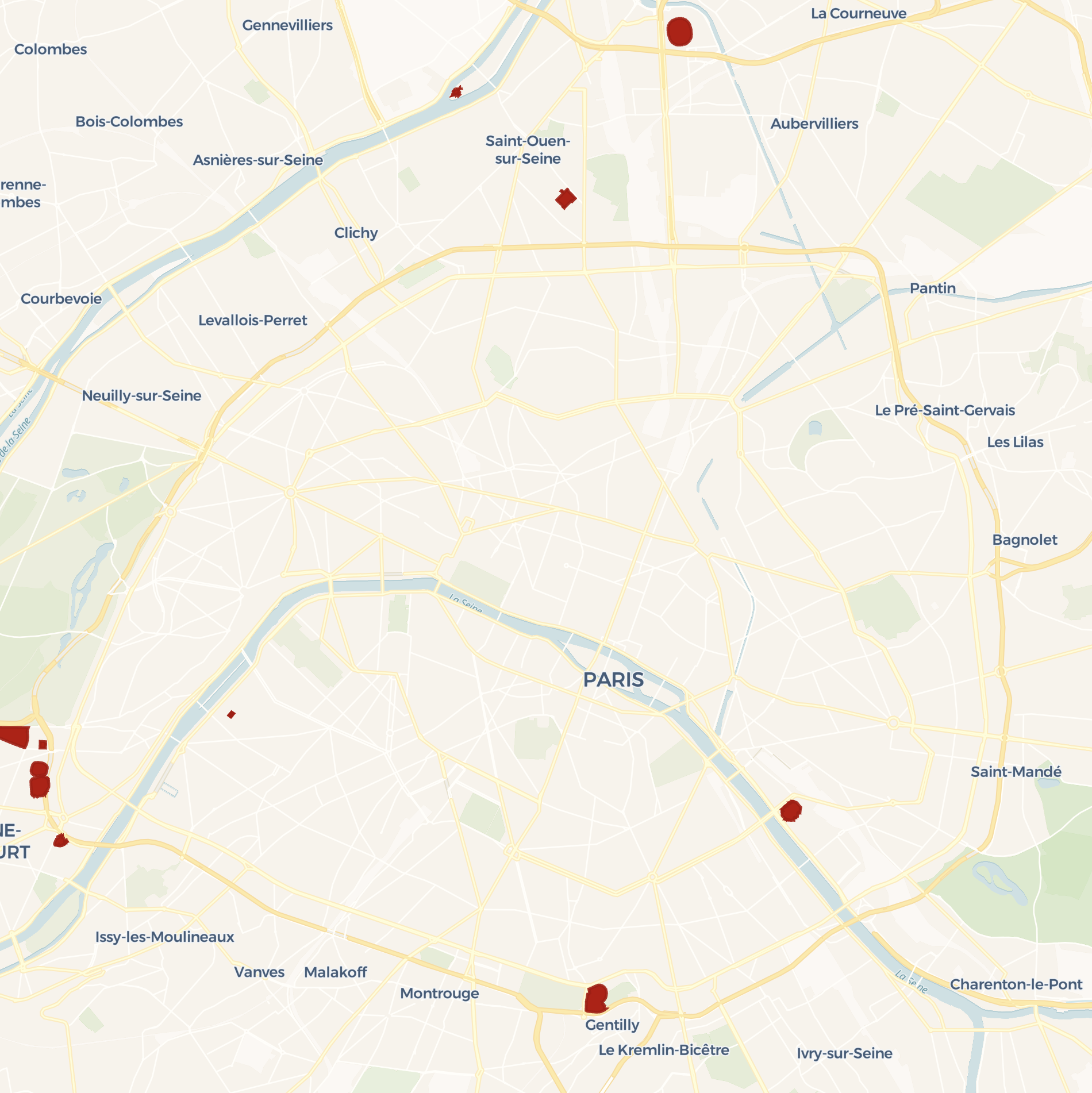}}
        \caption{Stadium buildings}
    \end{subfigure}
    \hfill
    \begin{subfigure}{0.35\linewidth}
        \includegraphics[width=\linewidth]{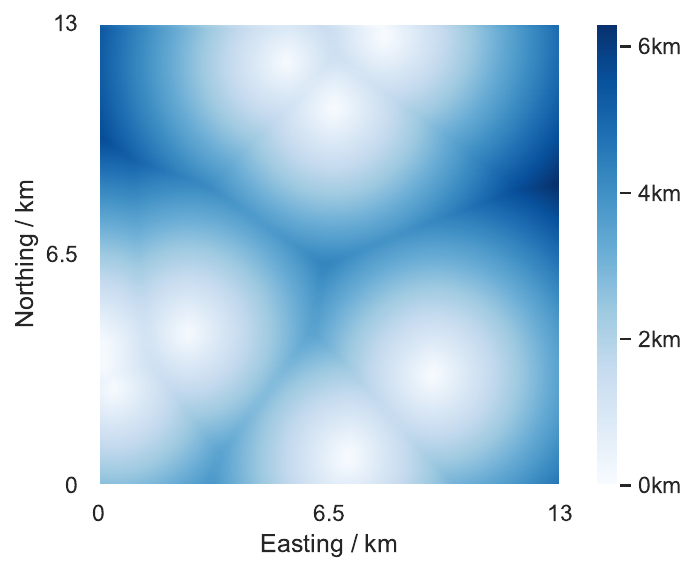}
        \caption{$\mu$ of \textit{distance(stadium)}}
    \end{subfigure}
    \hfill
    \begin{subfigure}{0.35\linewidth}
        \includegraphics[width=\linewidth]{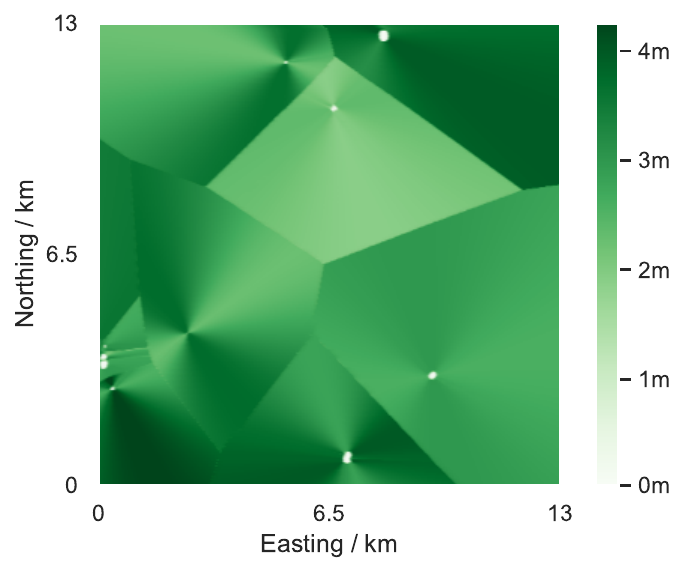}
        \caption{$\sigma$ of \textit{distance(stadium)}}
    \end{subfigure}\\
    \begin{subfigure}[b]{0.23\linewidth}
        \raisebox{5.75mm}{\includegraphics[width=\linewidth]{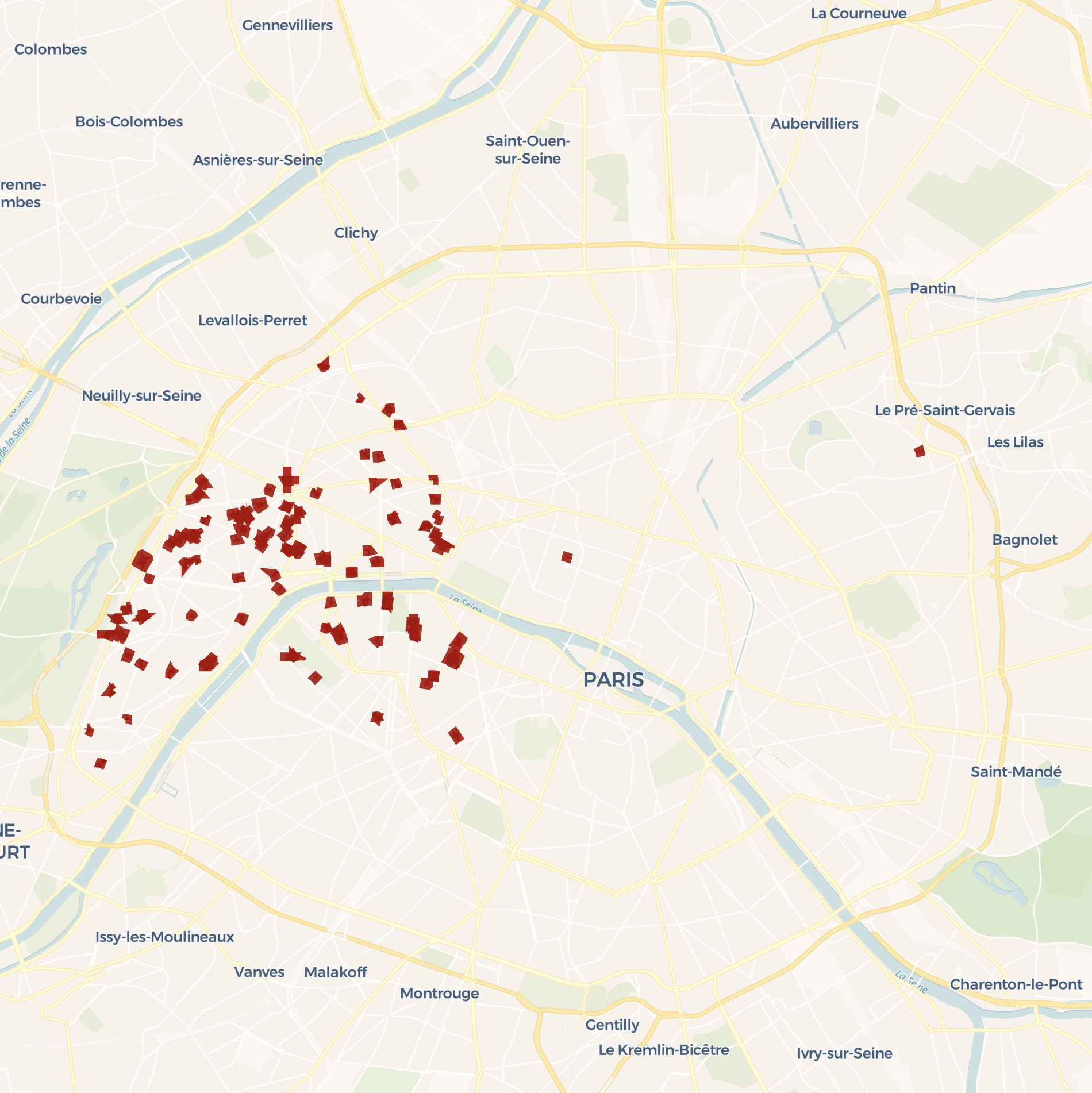}}
        \caption{Embassy buildings}
    \end{subfigure}
    \hfill
    \begin{subfigure}{0.35\linewidth}
        \includegraphics[width=\linewidth]{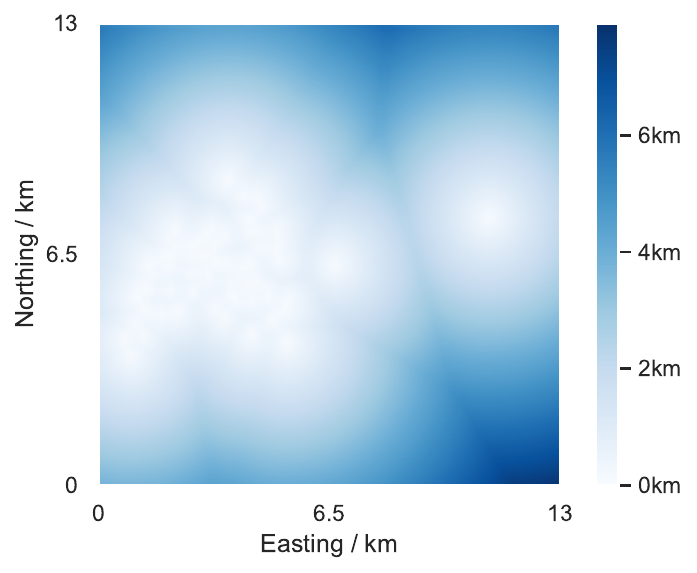}
        \caption{$\mu$ of \textit{distance(embassy)}}
    \end{subfigure}
    \hfill
    \begin{subfigure}{0.35\linewidth}
        \includegraphics[width=\linewidth]{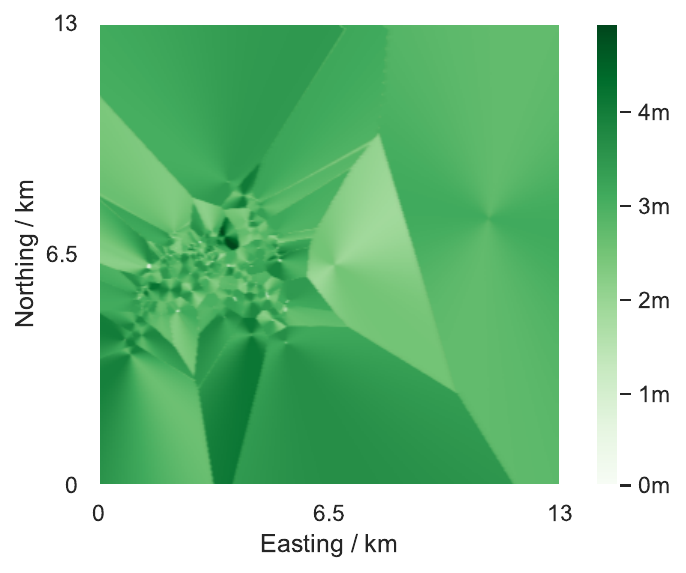}
        \caption{$\sigma$ of \textit{distance(embassy)}}
    \end{subfigure}\\
    \begin{subfigure}[b]{0.23\linewidth}
        \raisebox{5.75mm}{\includegraphics[width=\linewidth]{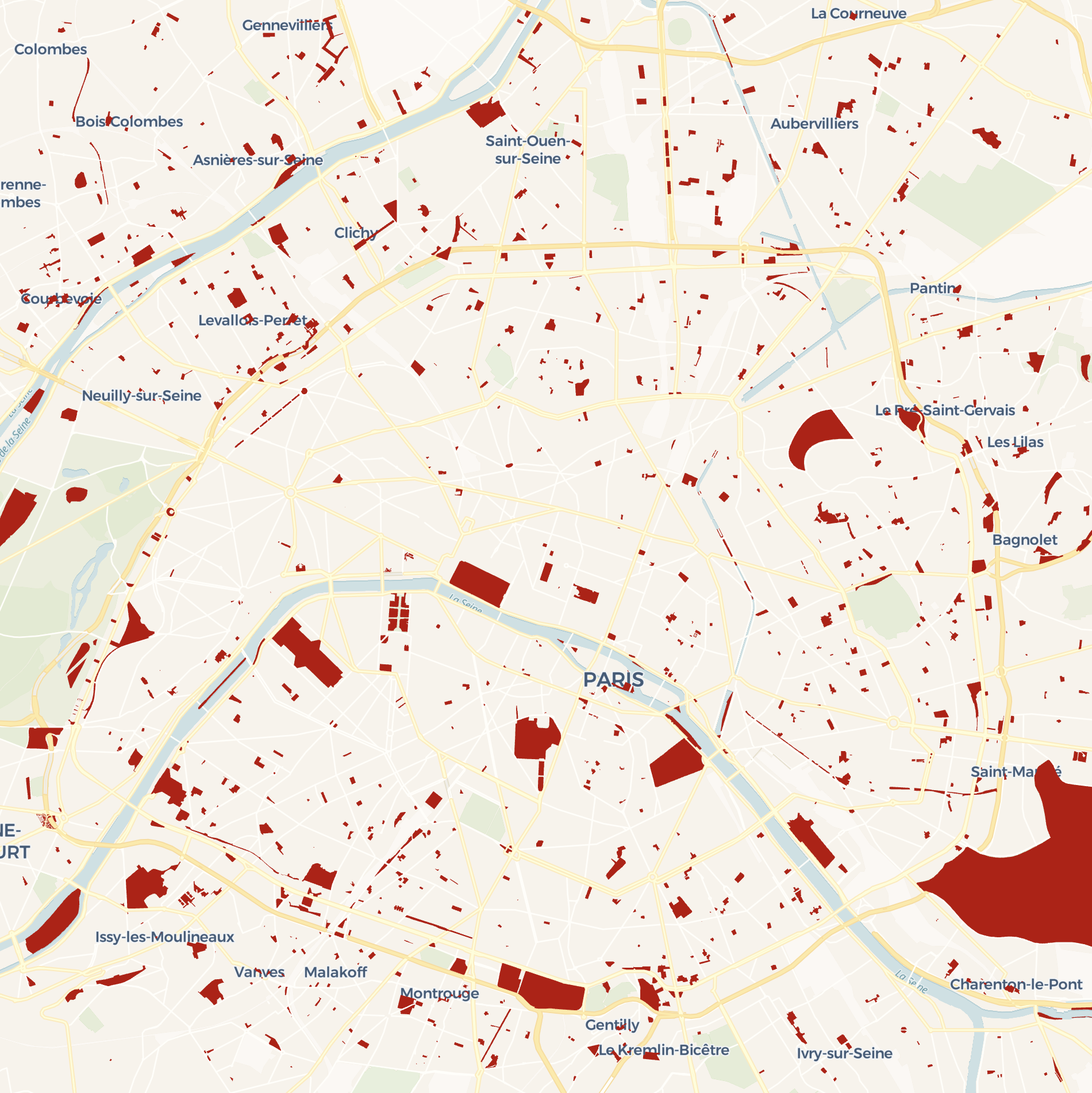}}
        \caption{Park areas}
    \end{subfigure}
    \hfill
    \begin{subfigure}{0.35\linewidth}
        \includegraphics[width=\linewidth]{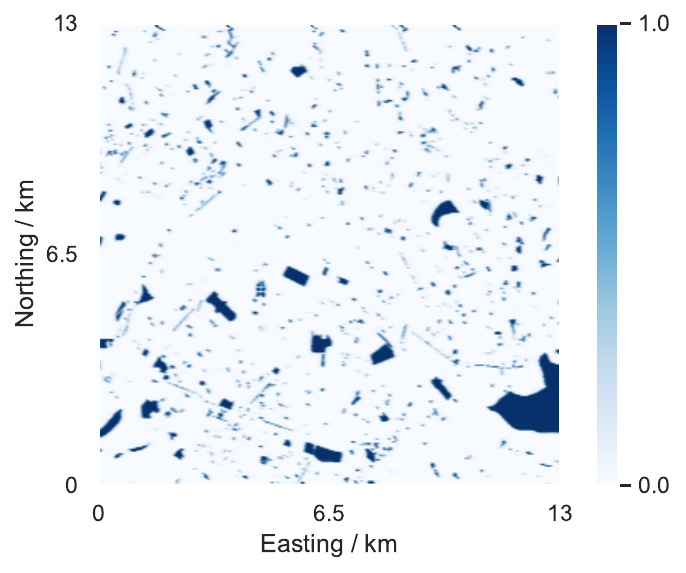}
        \caption{$\mu$ of \textit{over(park)}}
    \end{subfigure}
    \hfill
    \begin{subfigure}{0.35\linewidth}
        \includegraphics[width=\linewidth]{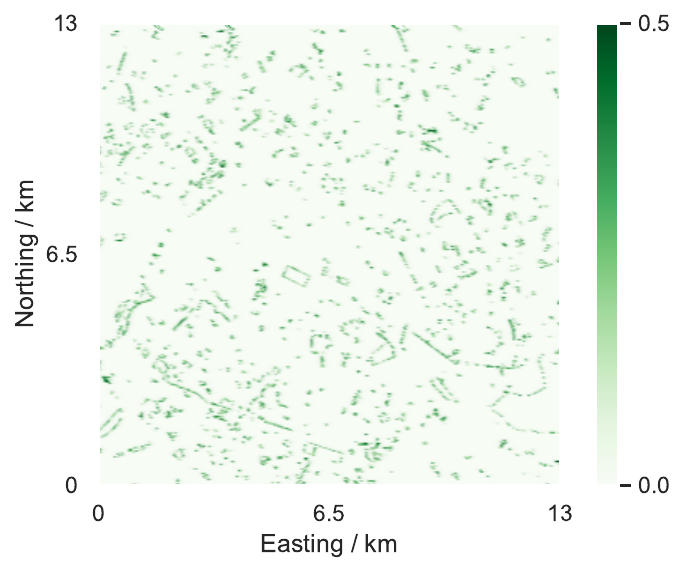}
        \caption{$\sigma$ of \textit{over(park)}}
    \end{subfigure}
    \caption{
        \textbf{Probabilistic spatial relations over Paris:}
        We visualize the source data and parameters employed for the probabilistic spatial relations used in Listing~\ref{listing:paris_constitution}.
        While the left column shows the relevant geometry, the center and right columns visualize the computed parameters of the respective relation.
    }
    \label{fig:paris_features}
\end{figure}

\begin{figure}
    \begin{subfigure}{0.32\linewidth}
        \includegraphics[width=\linewidth]{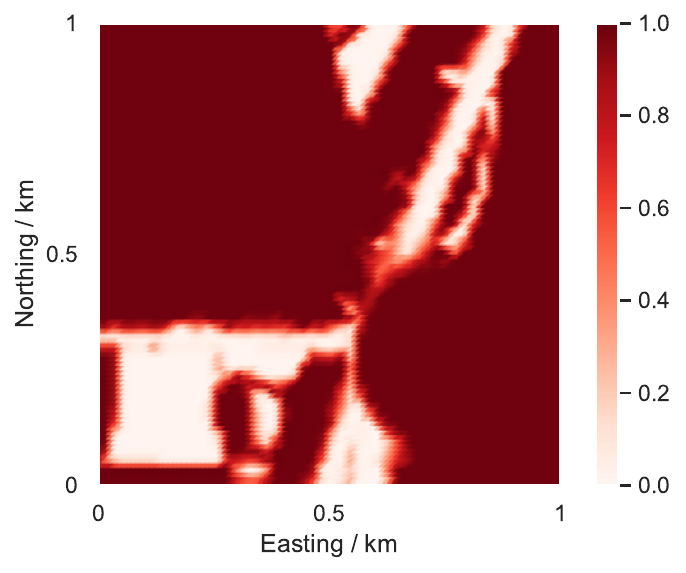}
        \caption{$\SI{50}{\meter}$ altitude}
    \end{subfigure}
    \hfill
    \begin{subfigure}{0.32\linewidth}
        \includegraphics[width=\linewidth]{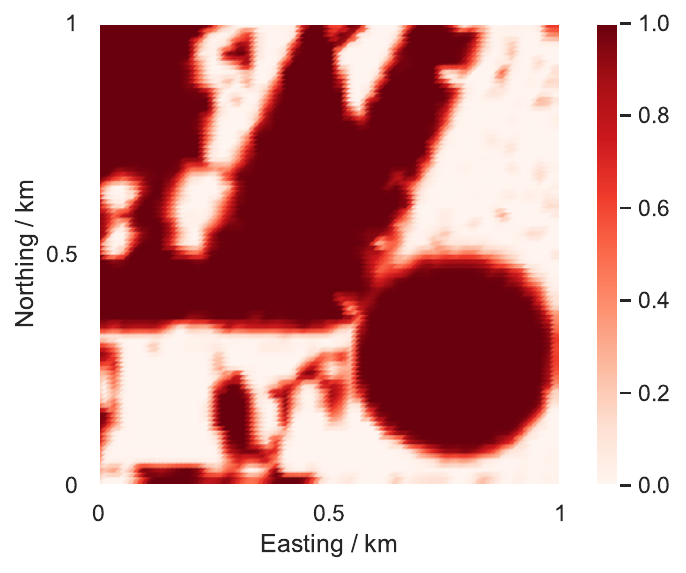}
        \caption{$\SI{150}{\meter}$ altitude}
    \end{subfigure}
    \hfill
    \begin{subfigure}{0.32\linewidth}
        \includegraphics[width=\linewidth]{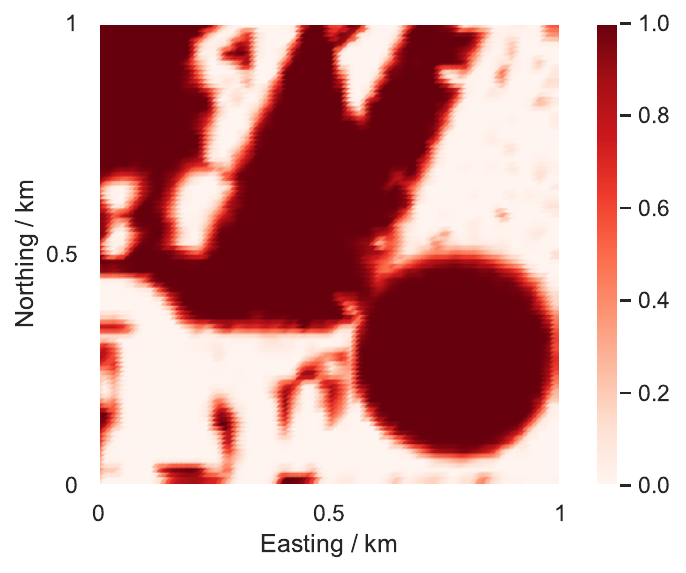}
        \caption{$\SI{250}{\meter}$ altitude}
    \end{subfigure} \\
    \begin{subfigure}{0.32\linewidth}
        \includegraphics[width=\linewidth]{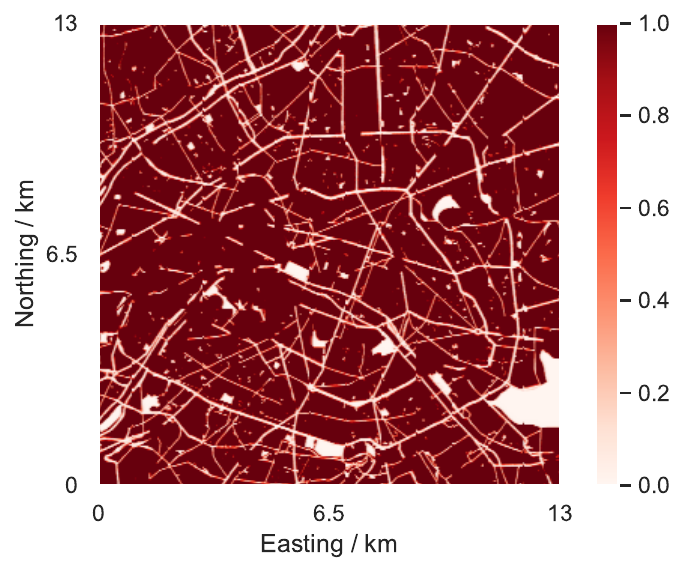}
        \caption{$\SI{50}{\meter}$ altitude}
    \end{subfigure}
    \hfill
    \begin{subfigure}{0.32\linewidth}
        \includegraphics[width=\linewidth]{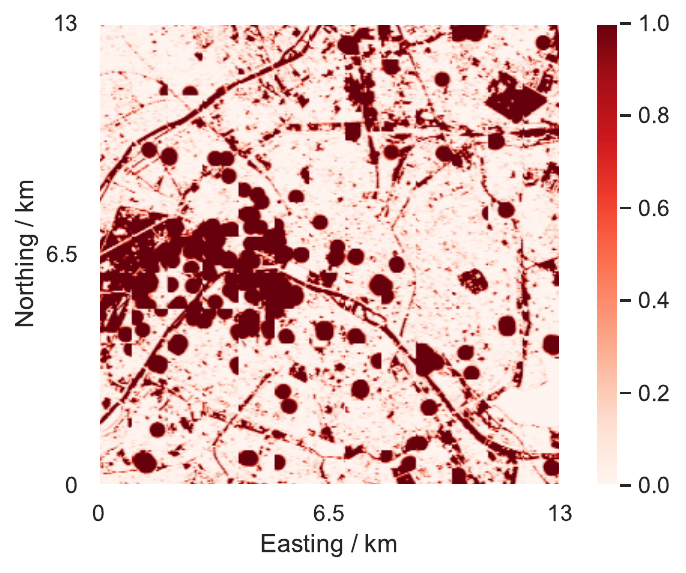}
        \caption{$\SI{150}{\meter}$ altitude}
    \end{subfigure}
    \hfill
    \begin{subfigure}{0.32\linewidth}
        \includegraphics[width=\linewidth]{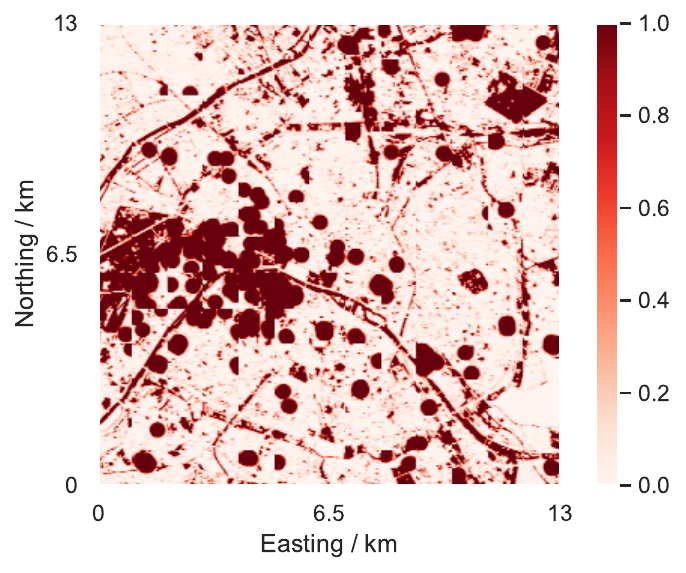}
        \caption{$\SI{250}{\meter}$ altitude}
    \end{subfigure}
    
    \caption{
        \textbf{Raising social acceptance by modeling legal requirements in hybrid probabilistic logic:}
        By querying the Constitution in Listing~\ref{listing:paris_constitution}, we obtain the probability of satisfying all proposed rules across Paris as shown in varying altitudes and scopes.
        By entering different altitude ranges in this scenario, the applicable set of rules changes, and here, larger volumes become accessible for the UAV given the choice of mission parameters (flight at \textit{daytime} with a \textit{standard\_license}).
    }
    \label{fig:paris_legality}
\end{figure}

From OpenStreetMap, we obtain crowd-sourced, real-world data on Paris.
To simulate restrictions as they might be imposed on an AAM scenario in urban areas, we consider relations about primary and secondary roads, buildings, sports stadiums, as well as government offices and embassies.
Hence, a variety of rules is created that may, in a similar form, be employed to create safe and socially acceptable AAM in densely populated urban areas.
Furthermore, in contrast to prior publications~\cite{kohaut2023mission, kohaut2024ceo}, we consider altitude-dependent rules for a full three-dimensional airspace regulation. 
Figure~\ref{fig:paris_features} shows all of the spatial relations and their respective parameters as have been employed for the next steps of UAV routing, probabilistic mission Clearance, Explanation, and Optimization.
We choose for the UAV routing setting the parameters \textit{daytime} and a \textit{standard\_license}, i.e., a comparatively restricted setting in comparison to, e.g., setting the \textit{expanded\_license} parameter instead.
The resulting landscapes of probabilities have been visualized in Figure~\ref{fig:paris_legality}.

\subsection{Physical Objectives}

We employ all physical objectives as discussed in Section~\ref{sec:physical_objectives}.
For each, we visualize the respective cost function at varying altitudes and scopes, once for a $\SI{1}{\kilo\meter\squared}$ tile and once for the entire $\SI{169}{\kilo\meter\squared}$ urban area.

\begin{figure}[ht!]
    \begin{subfigure}{0.32\linewidth}
        \includegraphics[width=\linewidth]{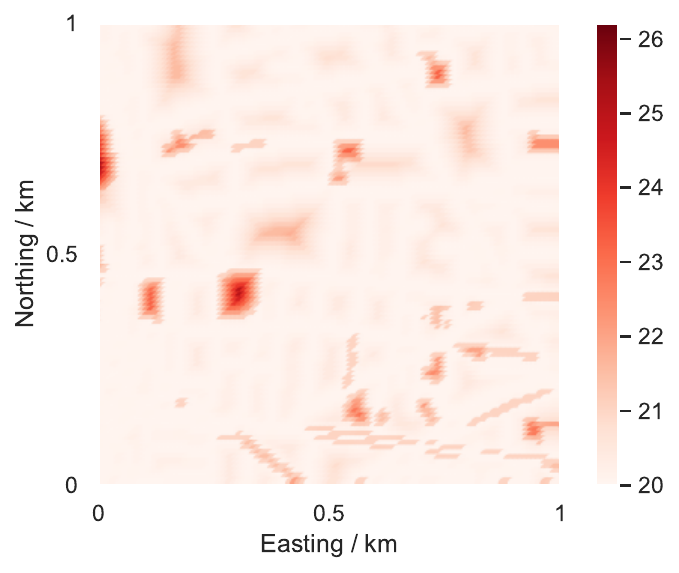}
        \caption{$\SI{50}{\meter}$ altitude}
    \end{subfigure}
    \hfill
    \begin{subfigure}{0.32\linewidth}
        \includegraphics[width=\linewidth]{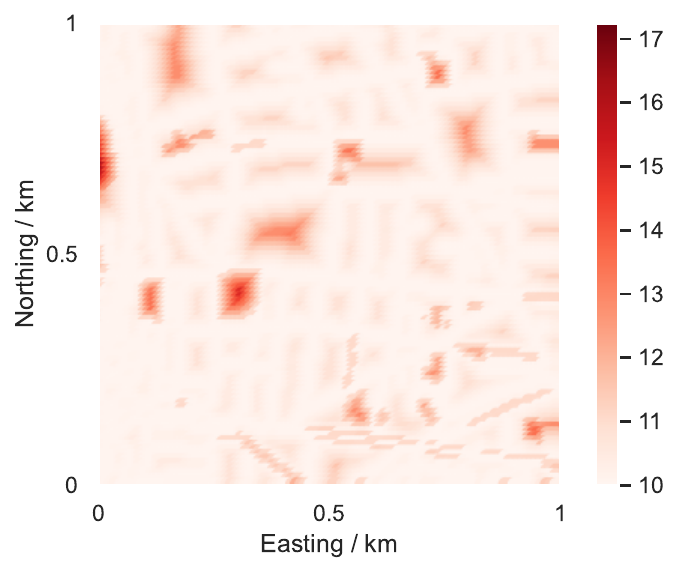}
        \caption{$\SI{150}{\meter}$ altitude}
    \end{subfigure}
    \hfill
    \begin{subfigure}{0.32\linewidth}
        \includegraphics[width=\linewidth]{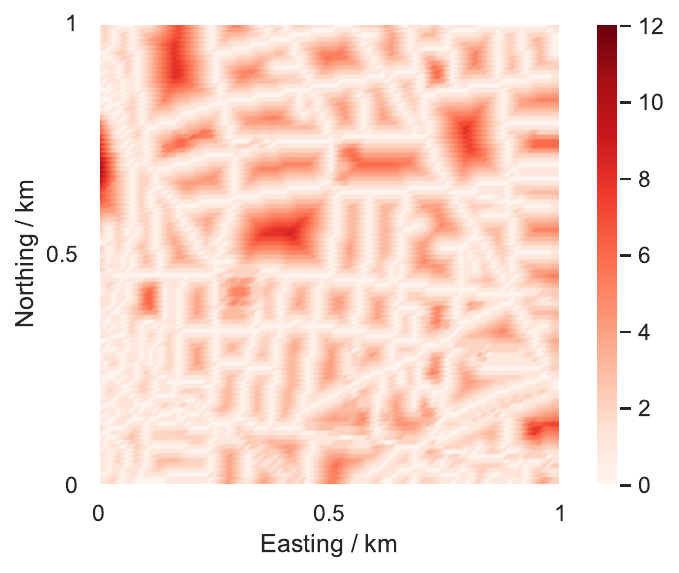}
        \caption{$\SI{250}{\meter}$ altitude}
    \end{subfigure} \\
    \begin{subfigure}{0.32\linewidth}
        \includegraphics[width=\linewidth]{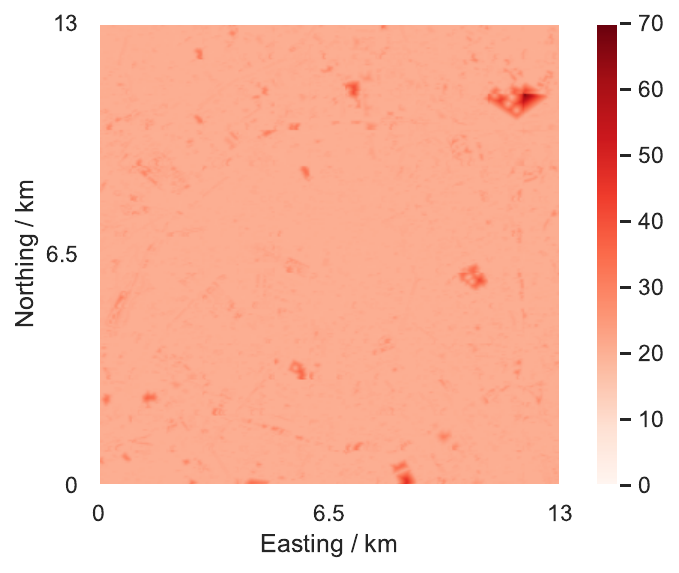}
        \caption{$\SI{50}{\meter}$ altitude}
    \end{subfigure}
    \hfill
    \begin{subfigure}{0.32\linewidth}
        \includegraphics[width=\linewidth]{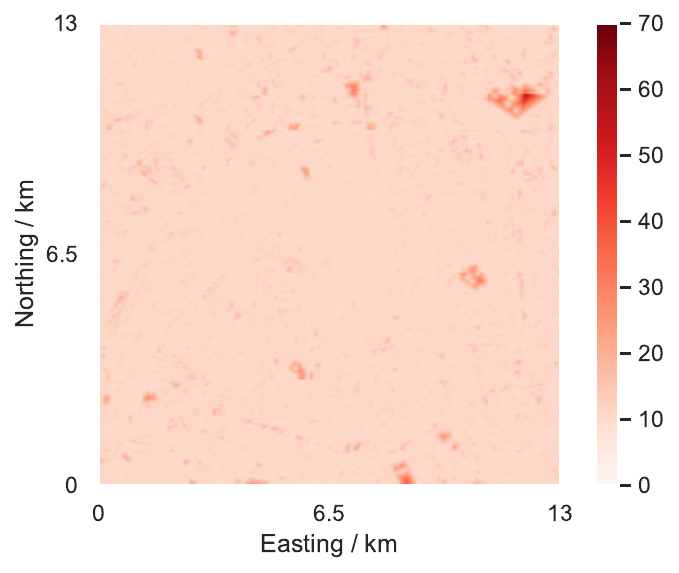}
        \caption{$\SI{150}{\meter}$ altitude}
    \end{subfigure}
    \hfill
    \begin{subfigure}{0.32\linewidth}
        \includegraphics[width=\linewidth]{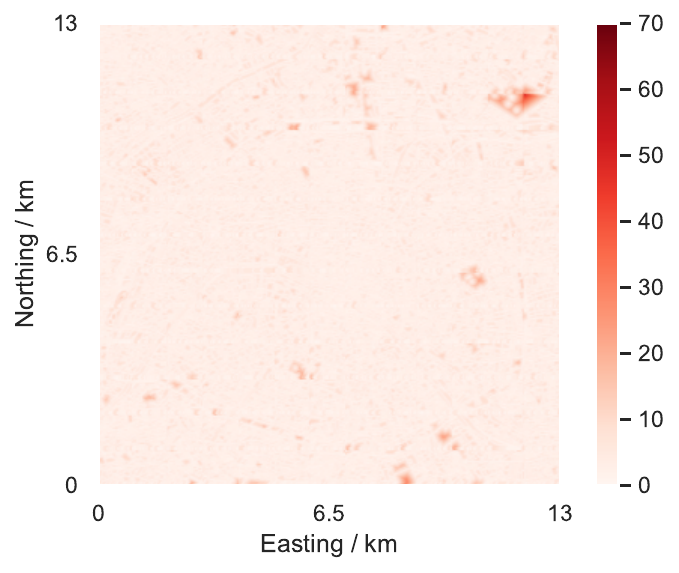}
        \caption{$\SI{250}{\meter}$ altitude}
    \end{subfigure}
    \caption{
        \textbf{Noise pollution as an objective for social acceptance.}
        \label{fig:paris_noise}
    }
\end{figure}

\begin{figure}[hb!]
    \begin{subfigure}{0.32\linewidth}
        \includegraphics[width=\linewidth]{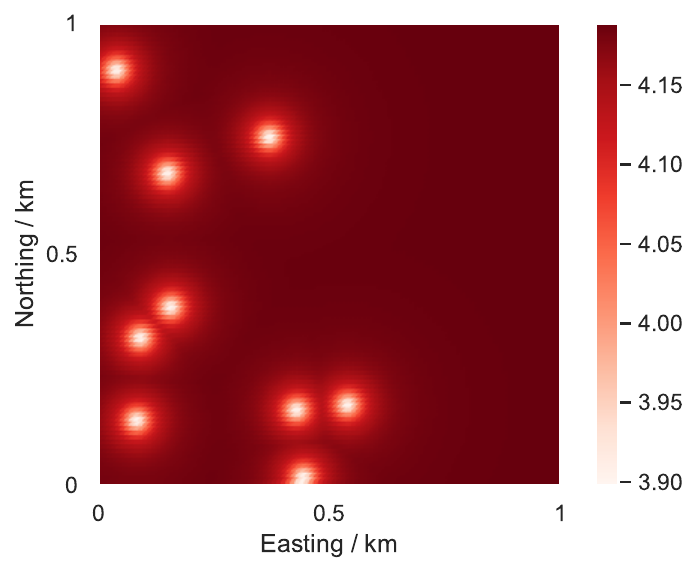}
        \caption{$\SI{50}{\meter}$ altitude}
    \end{subfigure}
    \hfill
    \begin{subfigure}{0.32\linewidth}
        \includegraphics[width=\linewidth]{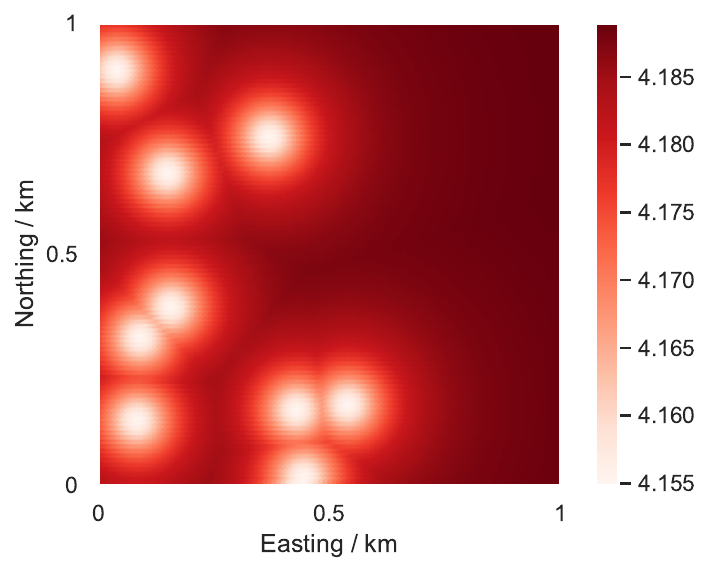}
        \caption{$\SI{150}{\meter}$ altitude}
    \end{subfigure}
    \hfill
    \begin{subfigure}{0.32\linewidth}
        \includegraphics[width=\linewidth]{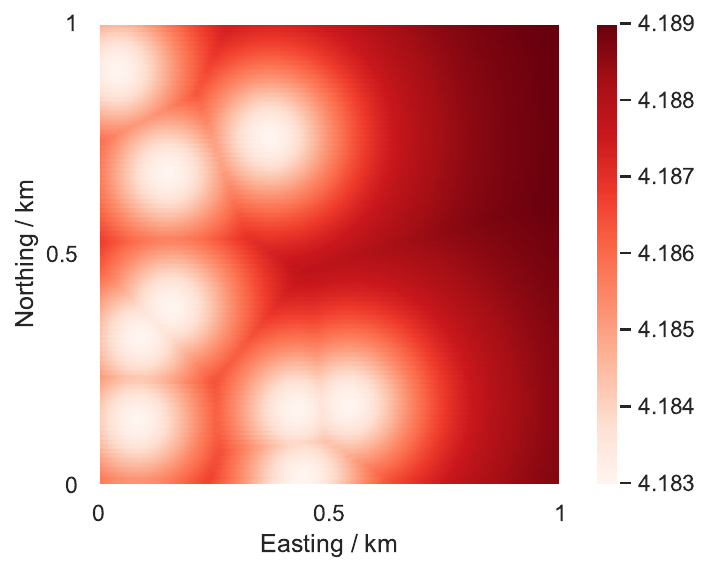}
        \caption{$\SI{250}{\meter}$ altitude}
    \end{subfigure} \\
    \begin{subfigure}{0.32\linewidth}
        \includegraphics[width=\linewidth]{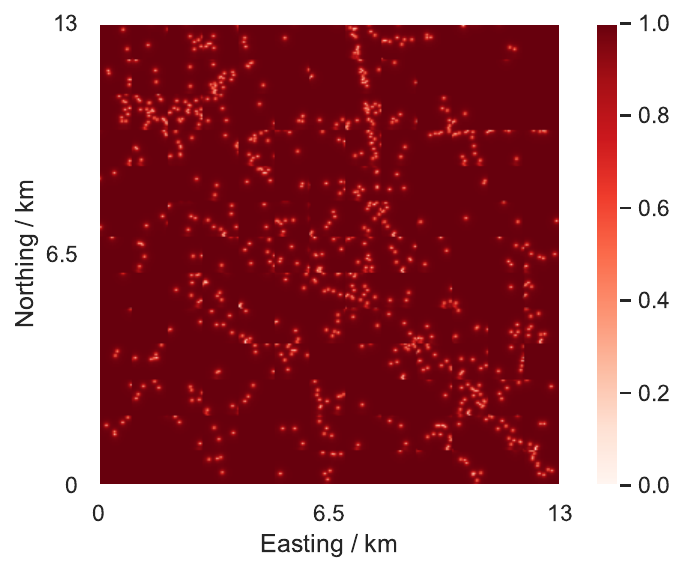}
        \caption{$\SI{50}{\meter}$ altitude}
    \end{subfigure}
    \hfill
    \begin{subfigure}{0.32\linewidth}
        \includegraphics[width=\linewidth]{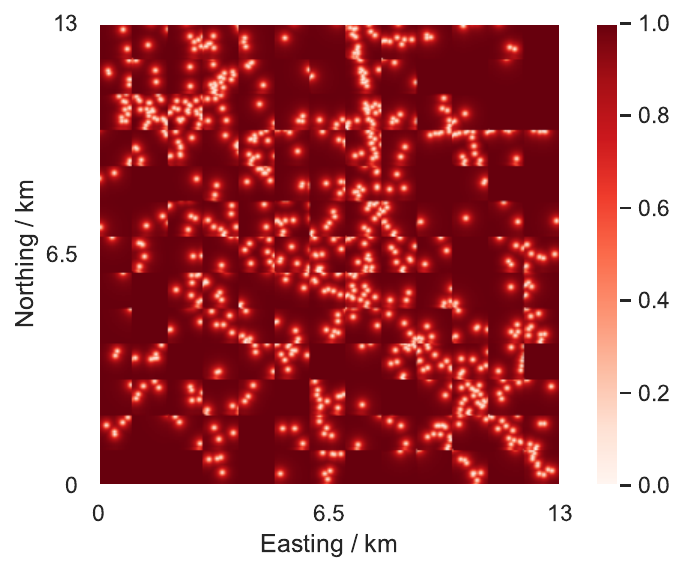}
        \caption{$\SI{150}{\meter}$ altitude}
    \end{subfigure}
    \hfill
    \begin{subfigure}{0.32\linewidth}
        \includegraphics[width=\linewidth]{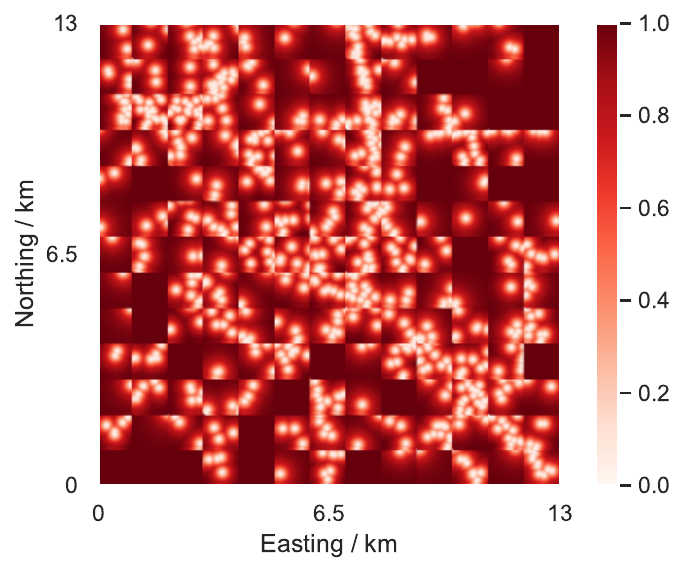}
        \caption{$\SI{250}{\meter}$ altitude}
    \end{subfigure}
    \caption{
        \textbf{Radio signal disturbance across Paris as modeled by Equation~\ref{eq:radio}.}
        \label{fig:paris_radio}
    }
\end{figure}

First, Figure~\ref{fig:paris_noise} visualizes the noise objective based on the geometry provided by OpenStreetMap.
As perceived noise disturbance lessens with an increased altitude of the UAV, so does the cost of the objective function.
Second, Figure~\ref{fig:paris_radio} shows the radio disturbance cost based on data provided by OpenCellID.
Here, it is advantageous for the UAV to stay close to at least one of the radio towers to maintain adequate signal quality.

\begin{figure}
    \begin{subfigure}{0.32\linewidth}
        \includegraphics[width=\linewidth]{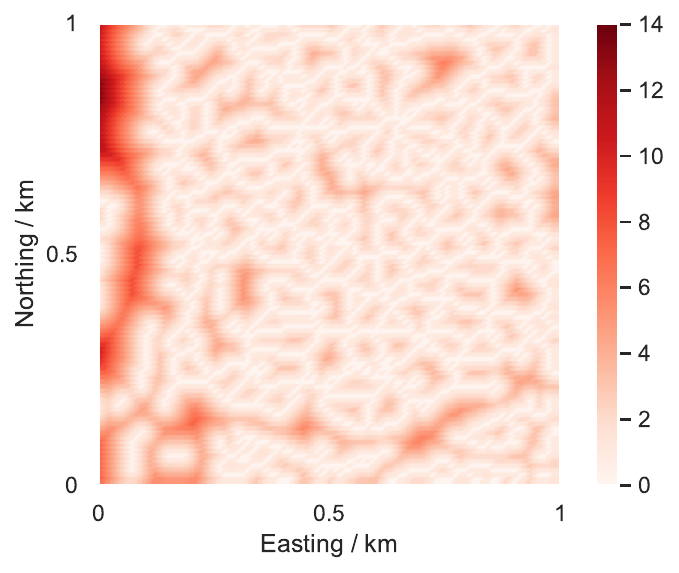}
        \caption{$\SI{50}{\meter}$ altitude}
    \end{subfigure}
    \hfill
    \begin{subfigure}{0.32\linewidth}
        \includegraphics[width=\linewidth]{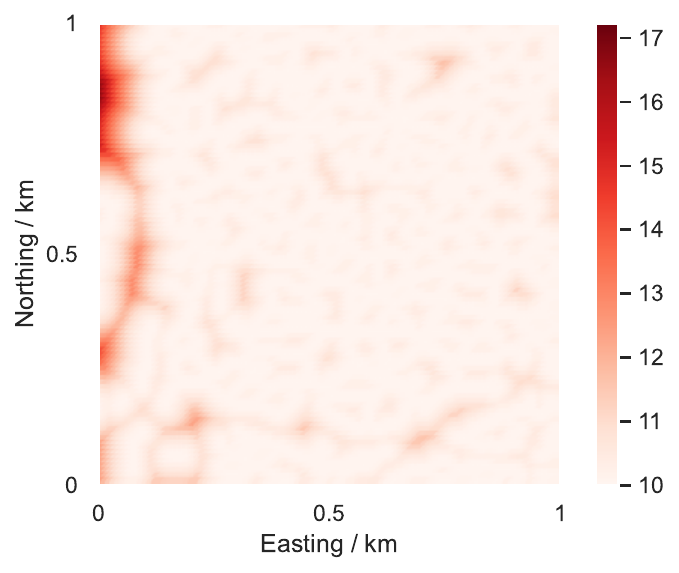}
        \caption{$\SI{150}{\meter}$ altitude}
    \end{subfigure}
    \hfill
    \begin{subfigure}{0.32\linewidth}
        \includegraphics[width=\linewidth]{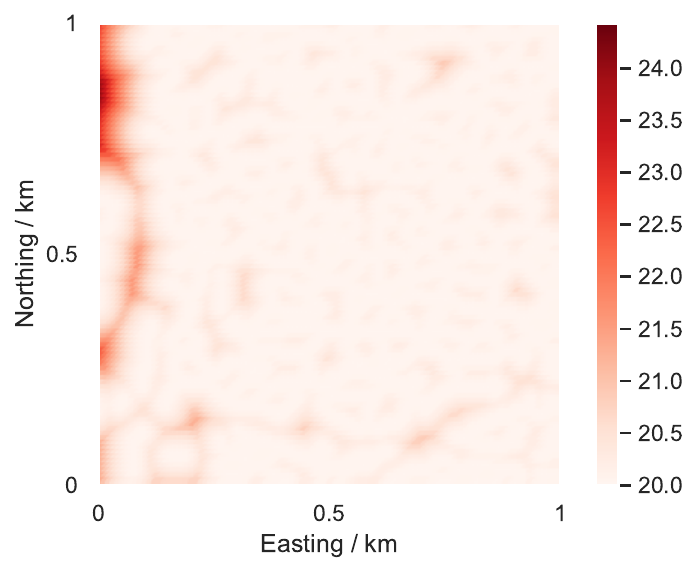}
        \caption{$\SI{250}{\meter}$ altitude}
    \end{subfigure} \\
    \begin{subfigure}{0.32\linewidth}
        \includegraphics[width=\linewidth]{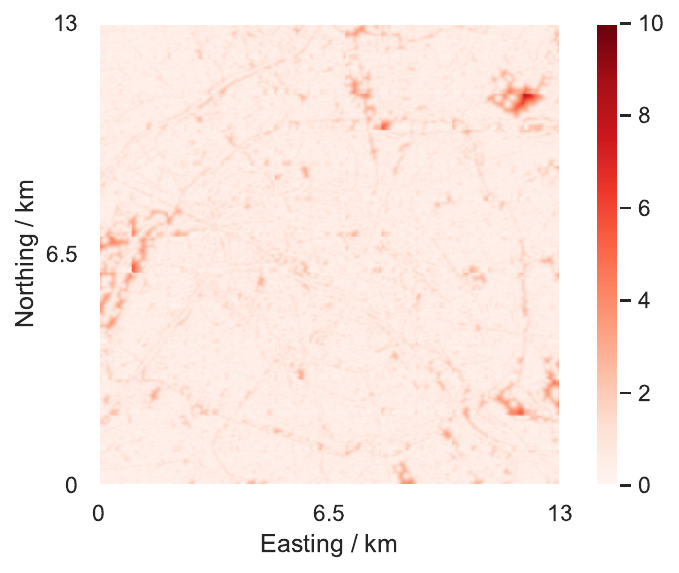}
        \caption{$\SI{50}{\meter}$ altitude}
    \end{subfigure}
    \hfill
    \begin{subfigure}{0.32\linewidth}
        \includegraphics[width=\linewidth]{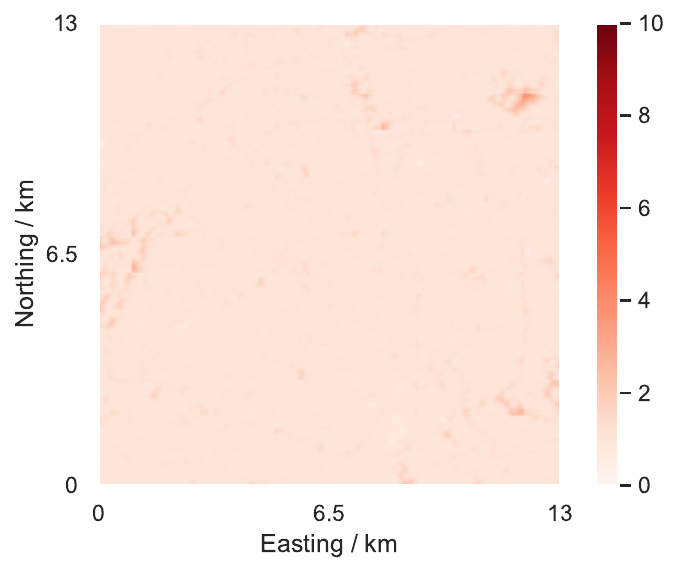}
        \caption{$\SI{150}{\meter}$ altitude}
    \end{subfigure}
    \hfill
    \begin{subfigure}{0.32\linewidth}
        \includegraphics[width=\linewidth]{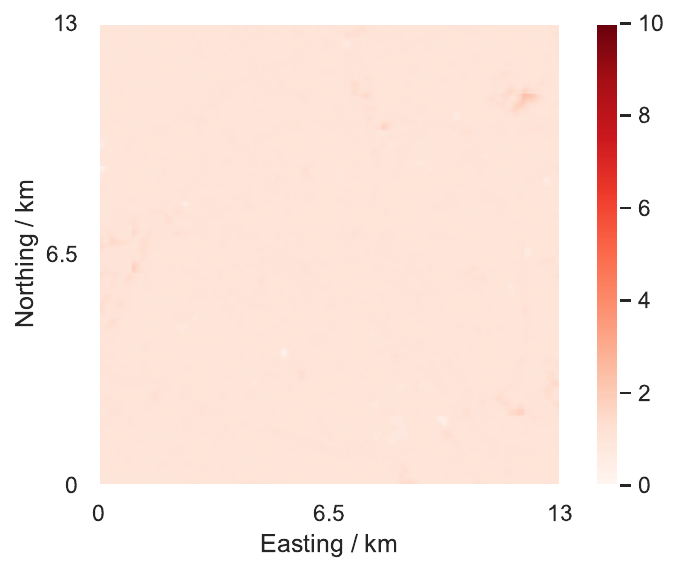}
        \caption{$\SI{250}{\meter}$ altitude}
    \end{subfigure}
    
    \caption{
        \textbf{Risk of injury objective across Paris.}
    }
    \label{fig:paris_risk}
\end{figure}

The signal quality degrades with the inverse square law as the distance of the UAV to the nearest tower increases.
For visualization purposes, the cost of radio disturbance has been normalized across the $\SI{1}{\km\squared}$ tiles in the Paris-wide overviews, with the discontinuities stemming from each tile loading the cell tower locations for planning in isolation.
Finally, Figure~\ref{fig:paris_risk} then illustrates the cost function based on the risk of injury.
Flying at low altitudes over buildings or water surfaces (compared to roads or parks) reduces this risk. 
Flying higher increases the risk, as the probability of the crash parabola ending over roads or parks increases. 
Therefore, the risk map becomes smoother at higher altitudes.


Together, these physical objectives form a basis for safe and effective UAV routing and, coupled with the compliance objective modeled in probabilistic first-order logic, are the input to the VIAS optimizer.

\subsection{Compliant and Effective UAV Routing}

\begin{figure}
    \begin{subfigure}{0.63\linewidth}
        \includegraphics[width=\linewidth]{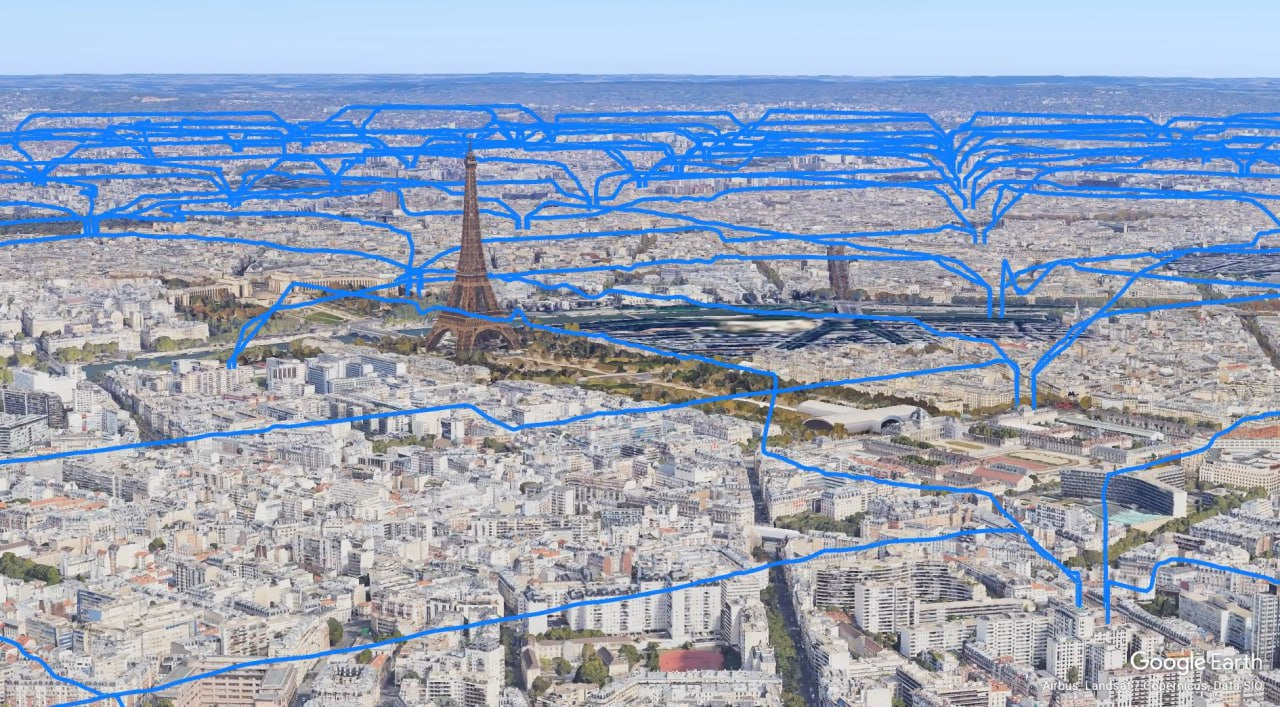}
        \caption{}
    \end{subfigure}
    \hfill
    \begin{subfigure}{0.35\linewidth}
        \includegraphics[width=\linewidth]{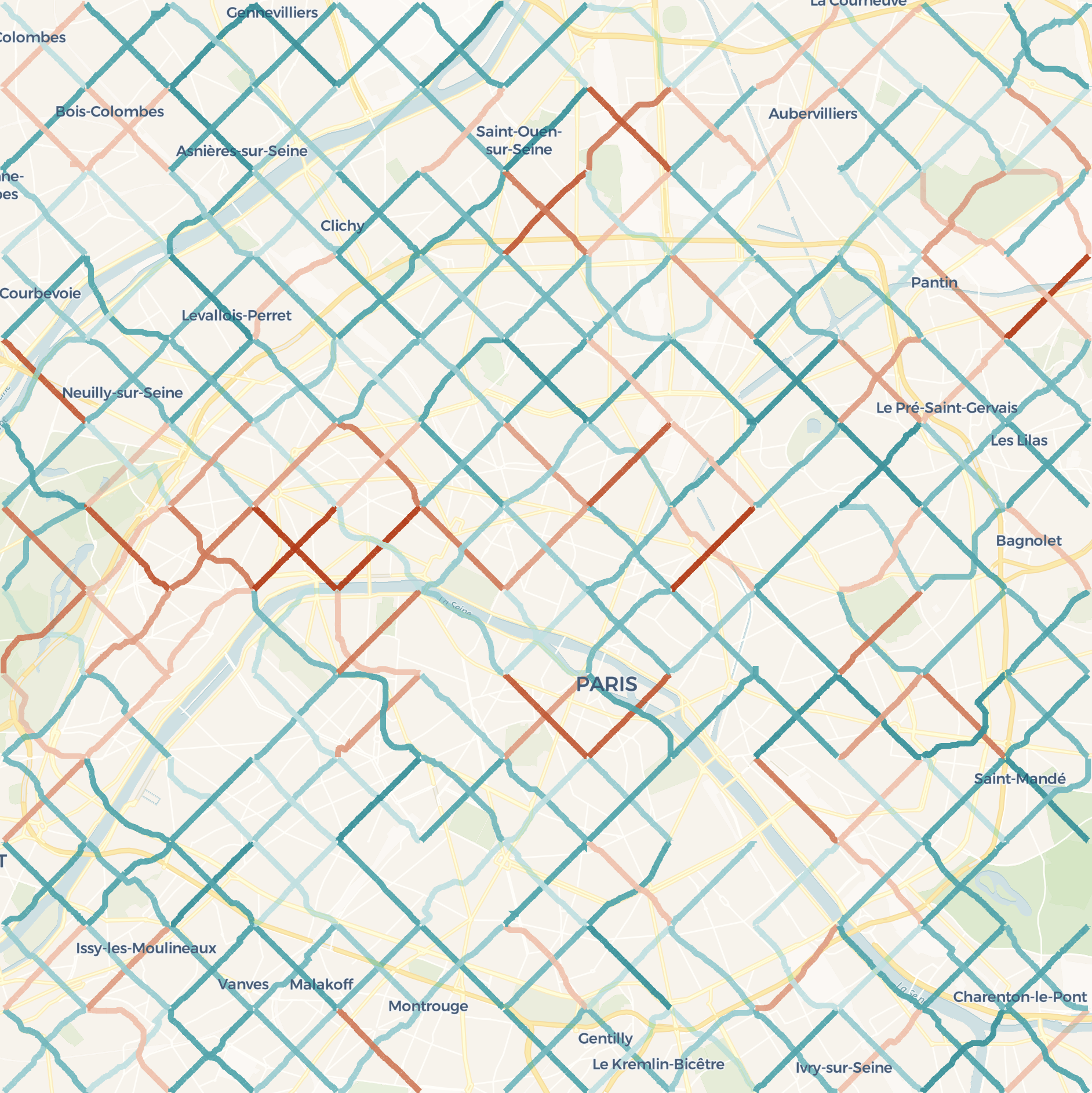}
        \caption{}
    \end{subfigure}

    \caption{
        \textbf{Knee point paths of the Pareto set:}
        Through the many-objective optimization, we obtain knee-point paths that sensibly trade off each of the employed physical and compliance objectives.
        This leads to paths that simultaneously receive high clearance scores (blue paths) while maintaining effectiveness, such as energy efficiency (straight paths).
    }
    \label{fig:paris_knee}
\end{figure}

Here, we show the resulting networks of paths from VIAS.
First, in Figure~\ref{fig:paris_knee}, we show the resulting UAV paths when choosing a sensible trade-off between all of the objectives: noise, radio disturbance, risk, energy, and compliance, respectively.
Hence, these paths achieve a balance between physical and social demands.
Here, (a) shows a 3D rendering~\cite{googleEarth} of the network over Paris around the Eiffel Tower, while (b) gives a bird's-eye overview of the entire covered area with paths drawn in blue if they are likely to satisfy the rules and red if they are likely to violate them.

\begin{figure}
    \begin{subfigure}{0.195\linewidth}
        \includegraphics[width=\linewidth]{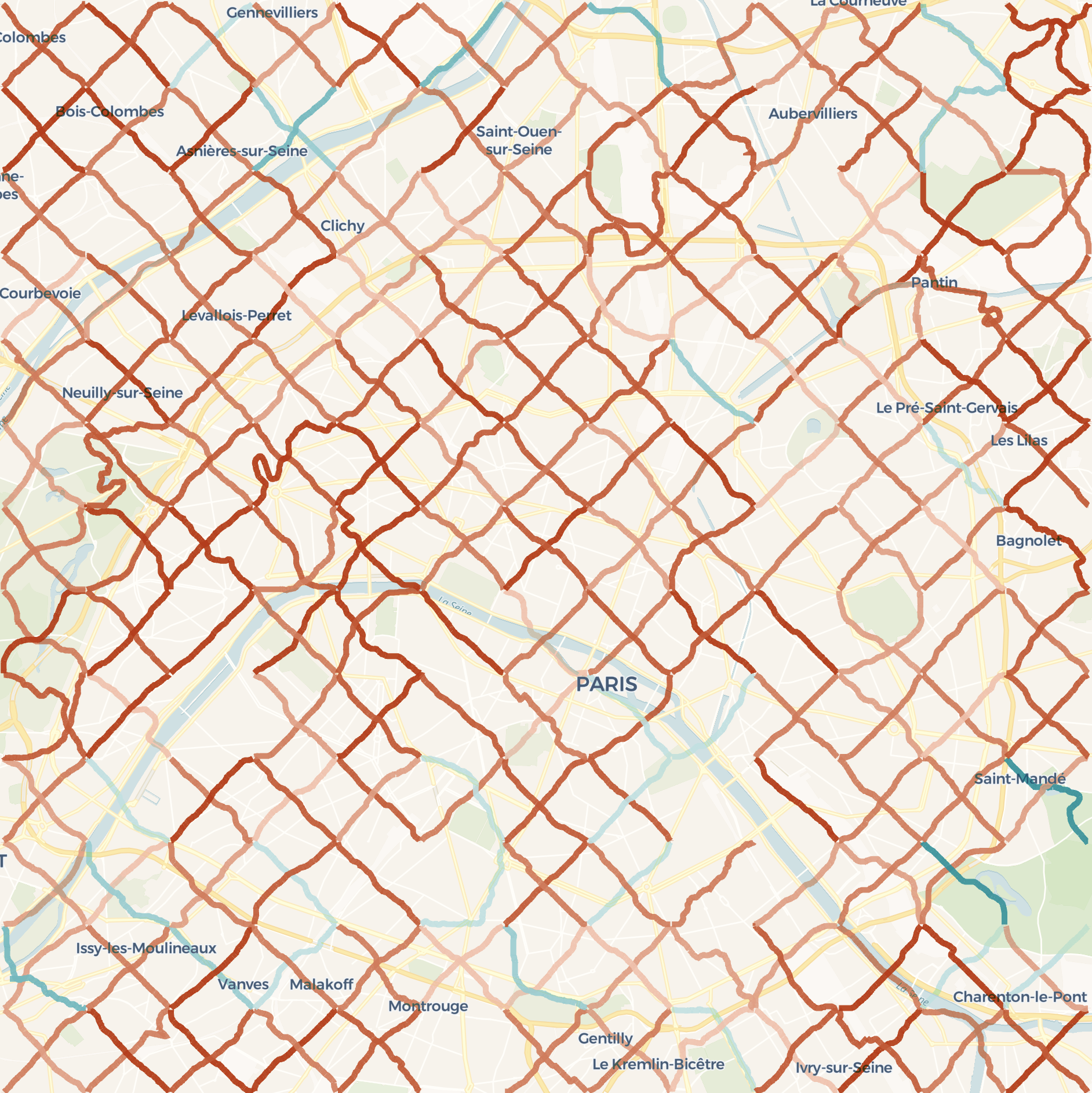}
        \caption{Risk}
    \end{subfigure}
    \hfill
    \begin{subfigure}{0.195\linewidth}
        \includegraphics[width=\linewidth]{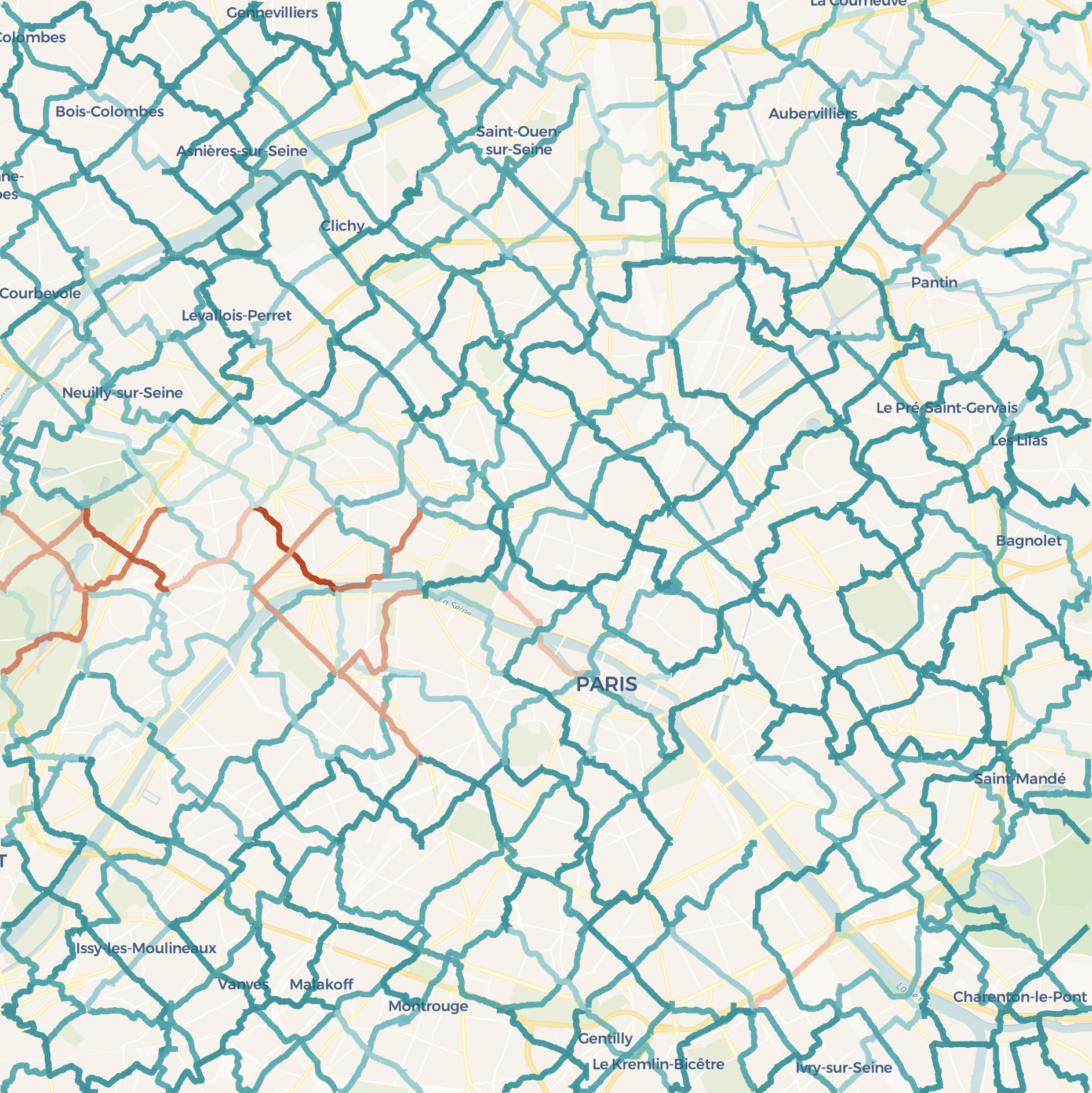}
        \caption{Noise}
    \end{subfigure}
    \hfill
    \begin{subfigure}{0.195\linewidth}
        \includegraphics[width=\linewidth]{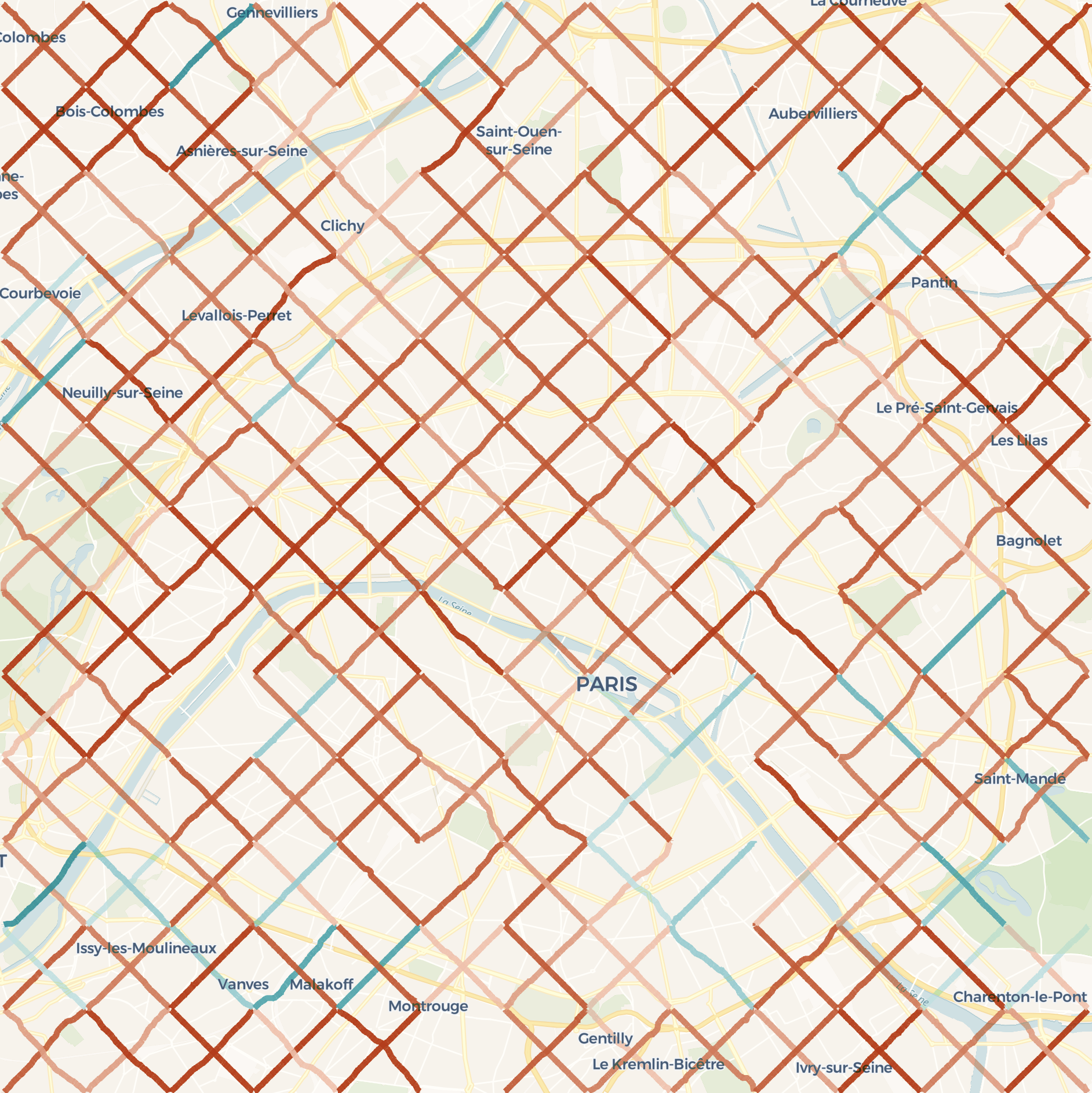}
        \caption{Radio disturbance}
    \end{subfigure}
    \hfill
    \begin{subfigure}{0.195\linewidth}
        \includegraphics[width=\linewidth]{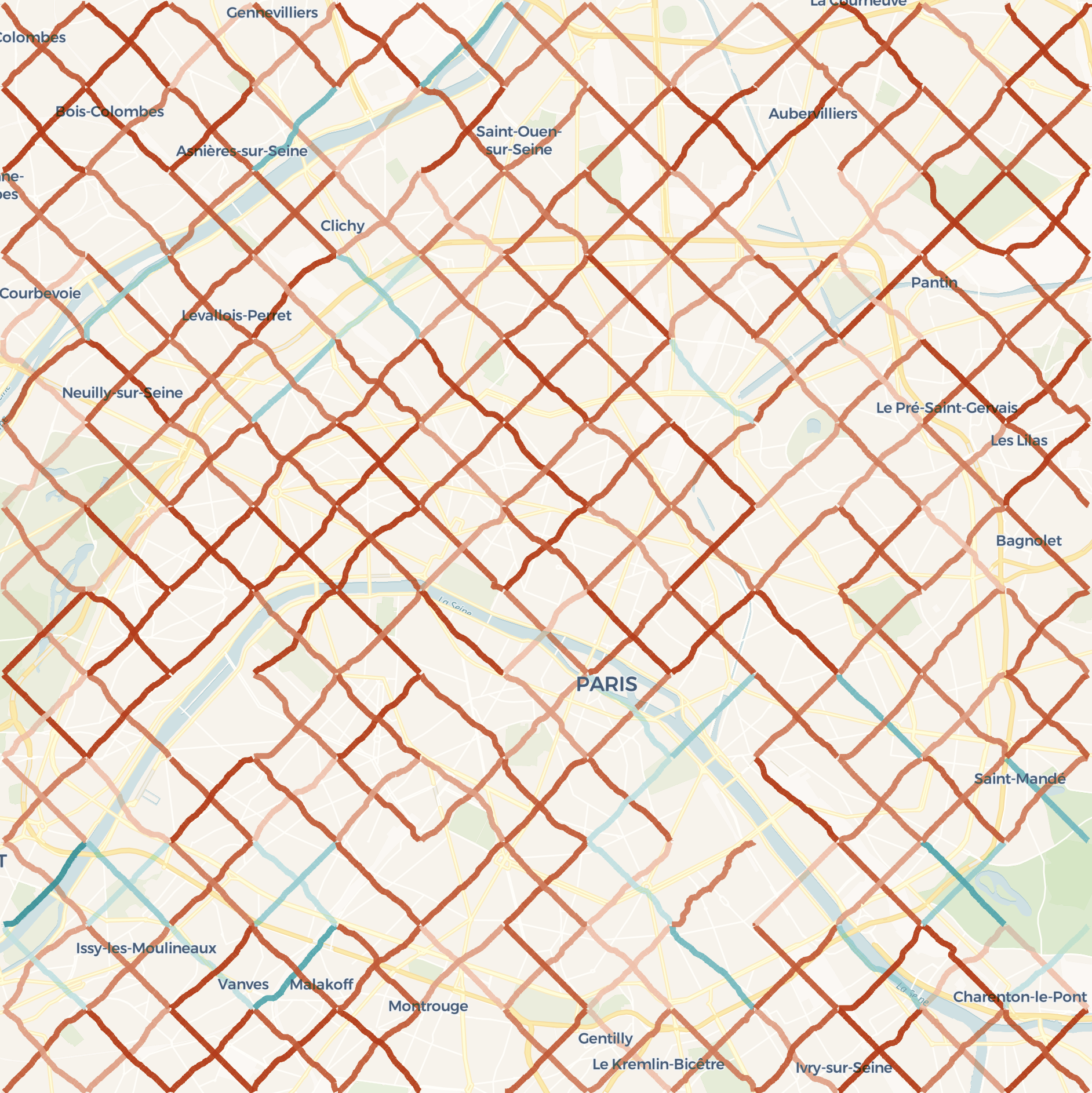}
        \caption{Energy}
    \end{subfigure}
    \hfill
    \begin{subfigure}{0.195\linewidth}
        \includegraphics[width=\linewidth]{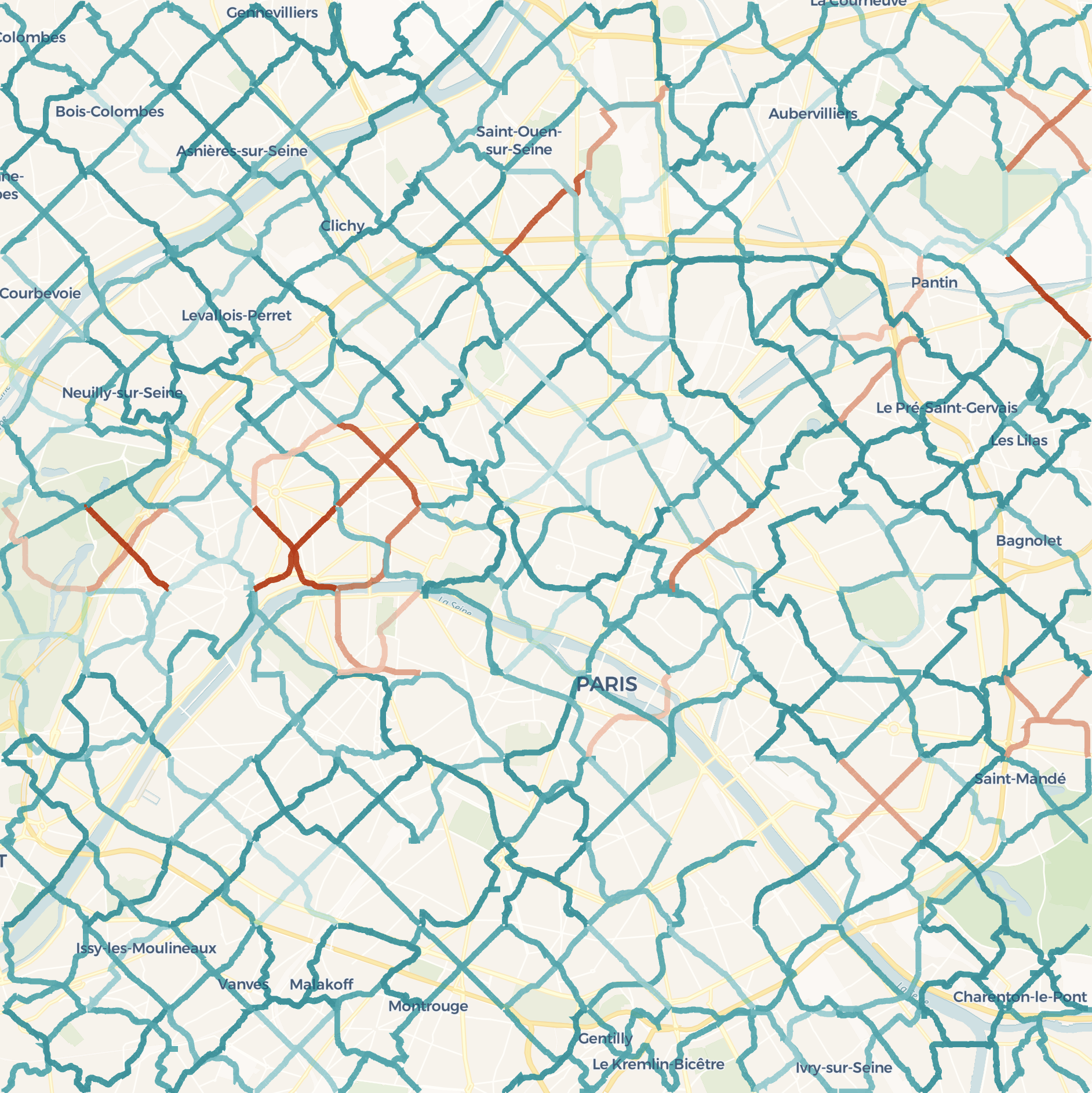}
        \caption{Compliance}
    \end{subfigure}

    \caption{
        \textbf{Extreme point paths of the Pareto sets:}
        By disregarding all other objectives, each individual objective may be optimized as much as possible.
        Hereby, e.g., a highly energy-efficient path may be obtained, but as no trade-off is applied, it is likely to be denied Clearance or stray too far away from the radio towers.
    }
    \label{fig:paris_extreme}
\end{figure}

Second, we visualize the extreme point paths in Figure~\ref{fig:paris_extreme}, showing the resulting UAV paths when choosing the extreme points for each objective: risk, noise, radio disturbance, energy, or compliance.
Hence, these paths mostly disregard every objective except one to match a single requirement best.
Furthermore, we show how likely each individual path is granted clearance by coloring them in blue for a clearance score as defined by Equation~\ref{eq:clearance} and red in case they obtain a low score and likely violate the rules along the way.

\begin{figure}
    \begin{subfigure}{0.345\linewidth}
        \includegraphics[width=\linewidth]{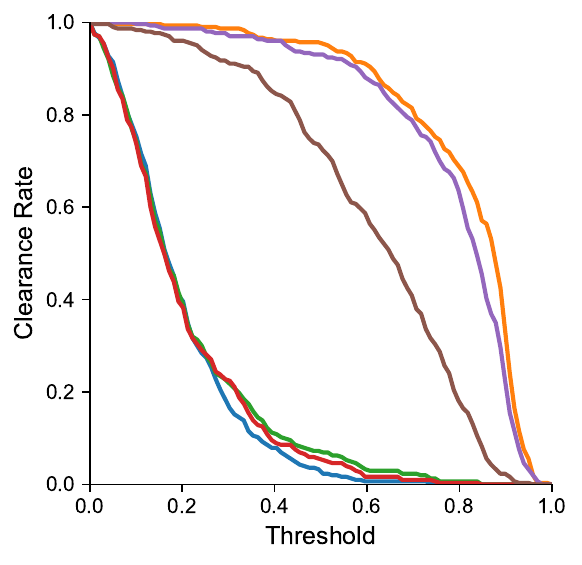}
        \caption{Standard license}
    \end{subfigure}
    \hfill
    \begin{subfigure}{0.345\linewidth}
        \includegraphics[width=\linewidth]{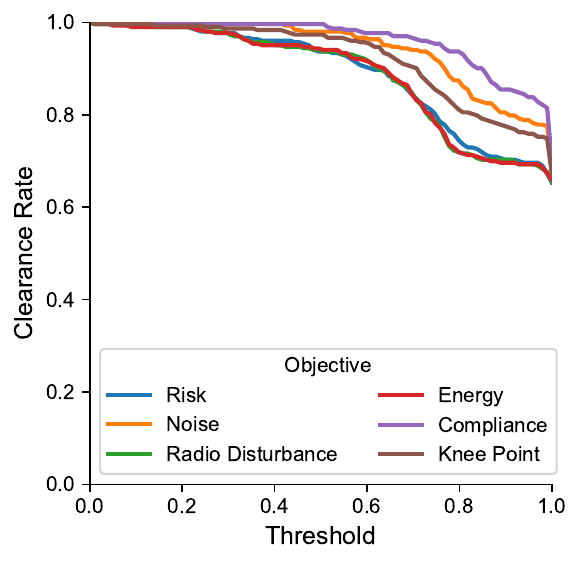}
        \caption{Expanded license}
    \end{subfigure}
    \hfill
    \begin{subfigure}[b]{0.29\linewidth}
        \raisebox{5.75mm}{\includegraphics[width=\linewidth]{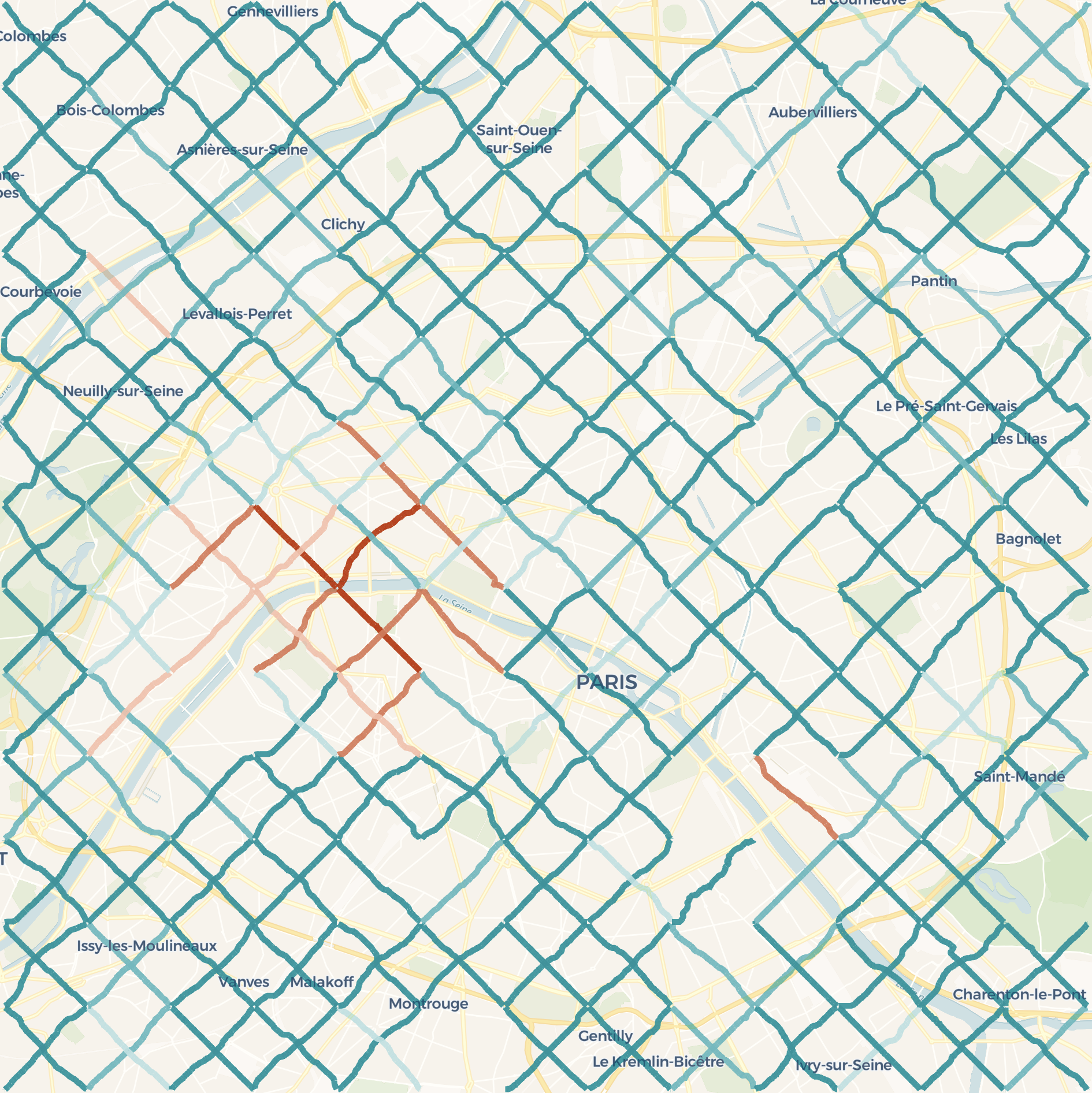}}
        \caption{Energy optimal paths}
    \end{subfigure}
    \caption{
        \textbf{Explanation and Optimization of proposal paths:}
        Through Mission Explanation and Optimization, the system may provide a setting in which even extreme point paths are likely to be granted Clearance.
        In this case, choosing the expanded license parameter leads to a significantly increased clearance rate.
        However, real-world limitations may apply, e.g., the suggested license may not be available to operators on-site.
    }
    \label{fig:explanation_optimization}
\end{figure}

Third, we demonstrate Explanation and Optimization as described in Sections~\ref{sec:explanation} and \ref{sec:optimization} in Figure~\ref{fig:explanation_optimization}.
Initially, the parameters in Listing~\ref{listing:paris_constitution} were chosen such that the UAV was assumed to be employed during daytime with a standard license and the respectively modeled restrictions, leading to many paths being denied clearance in both the knee-point (Figure~\ref{fig:paris_knee}) and extreme point paths (Figure~\ref{fig:paris_extreme}).
By searching the parameter space, we can automatically (i) explain that Clearance is mostly denied due to the choice in license and (ii) propose a more optimal mission setting by choosing the expanded license.

In Figure~\ref{fig:explanation_optimization}, we show the rate at which paths obtain clearance with varying thresholds for each extreme point and the knee point paths, once for the original setting (a) and once for the optimal setting (b).
In (c), one can see that even the energy extreme point paths receive mostly high clearance scores compared to Figure~\ref{fig:paris_extreme}, meaning that the framework can point the mission design towards settings that allow for satisfactory and highly effective routing even without specifically optimizing paths for compliance.

\begin{table}[]
    \caption{
        Path rejection areas for different choices of license parameter and objective prioritization.
        A smaller rejection area, as is the case when using the expanded license, implies a larger subset of the proposal paths may receive clearance for a non-zero threshold.
    }
    \begin{tabular}{clr}
        \toprule
        Mission Setting & Objective Priority & Area over Curve (Figure~\ref{fig:explanation_optimization} (a) and (b))\\ 
        \midrule 
        \multirow{6}{*}{\shortstack{Standard\\License}}
        & Risk optimal & 0.799 \\
        & Noise optimal & 0.188 \\
        & Radio Disturbance optimal & 0.784 \\
        & Energy optimal & 0.791 \\
        & Compliance optimal & 0.213 \\
        & Balanced (Knee Point) & 0.384 \\
        \midrule 
        \multirow{6}{*}{\shortstack{Expanded\\License}}
        & Risk optimal & 0.108 \\
        & Noise optimal & 0.056 \\
        & Radio Disturbance optimal & 0.112 \\
        & Energy optimal & 0.112 \\
        & Compliance optimal & 0.037 \\
        & Balanced (Knee Point) & 0.075 \\
        \bottomrule
    \end{tabular}
    \label{tab:rejection_areas}
\end{table}

\subsection{Runtime Requirements}

Our framework aims to provide compliant and effective UAV routing in complex and demanding environments such as urban cores.
Because this is a computationally demanding task that depends on the application environment, we discuss our framework's runtime and current bottlenecks.

\begin{figure}
    \begin{subfigure}{0.325\linewidth}
        \includegraphics[width=\linewidth]{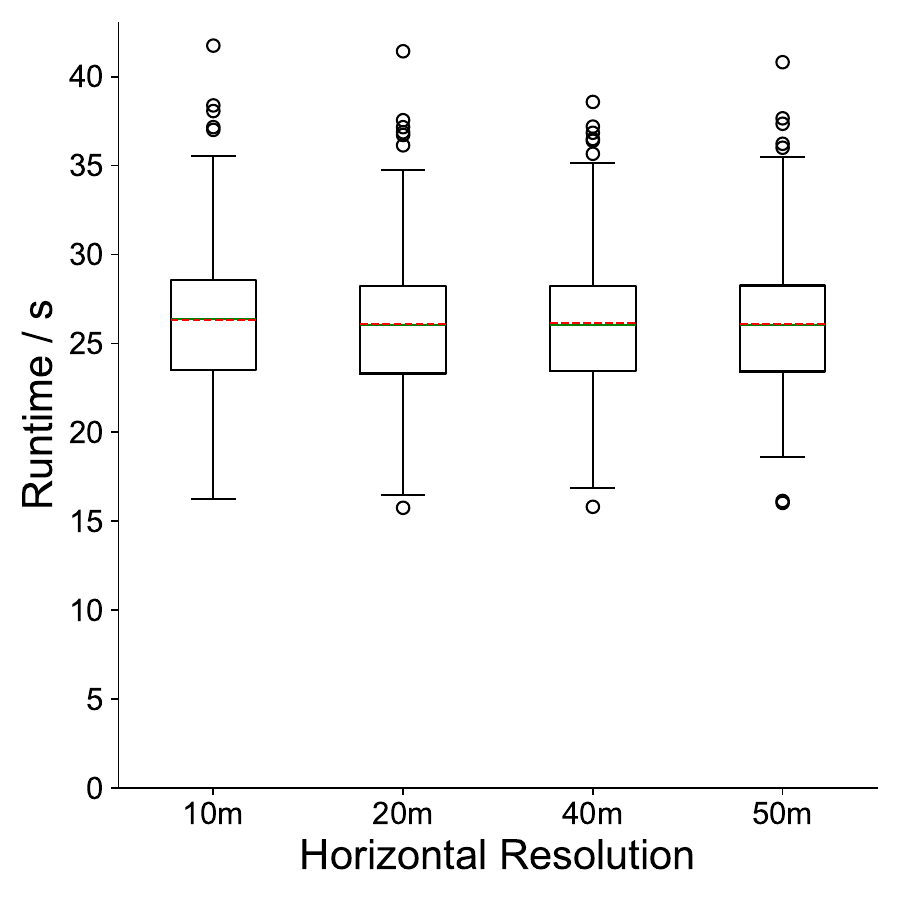}
        \caption{StaR Map Preparation}
    \end{subfigure}    
    \hfill
    \begin{subfigure}{0.325\linewidth}
        \includegraphics[width=\linewidth]{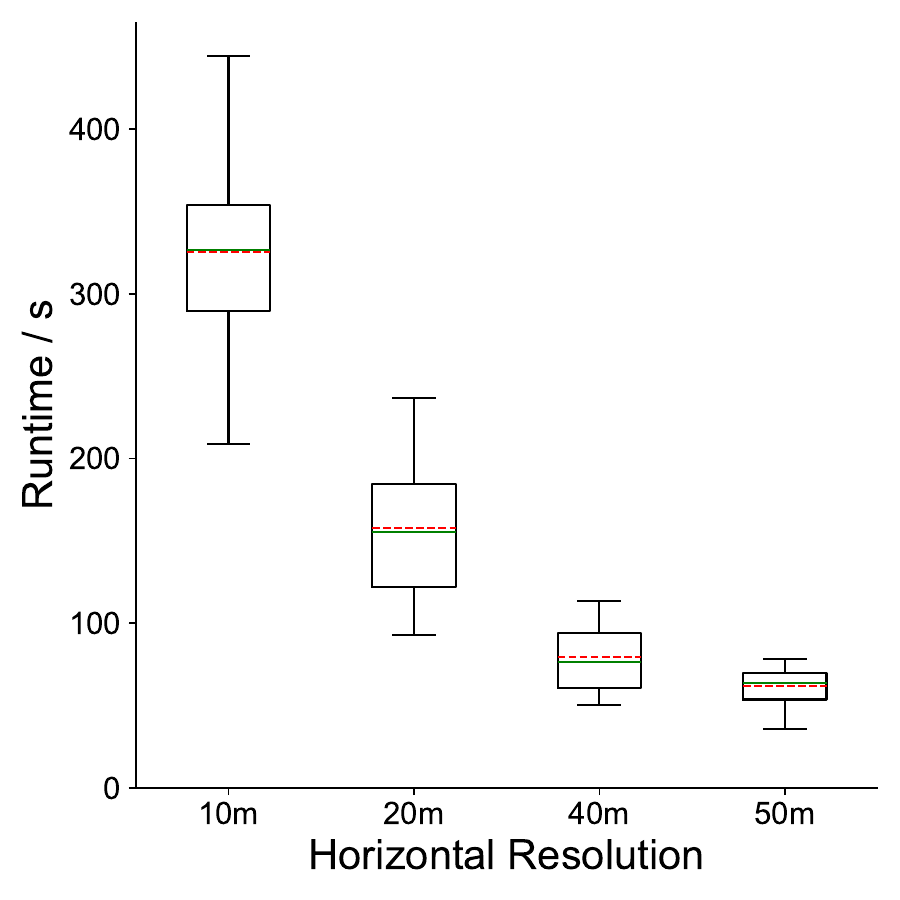}
        \caption{VIAS Preparation}
    \end{subfigure}
    \hfill
    \begin{subfigure}{0.325\linewidth}
        \includegraphics[width=\linewidth]{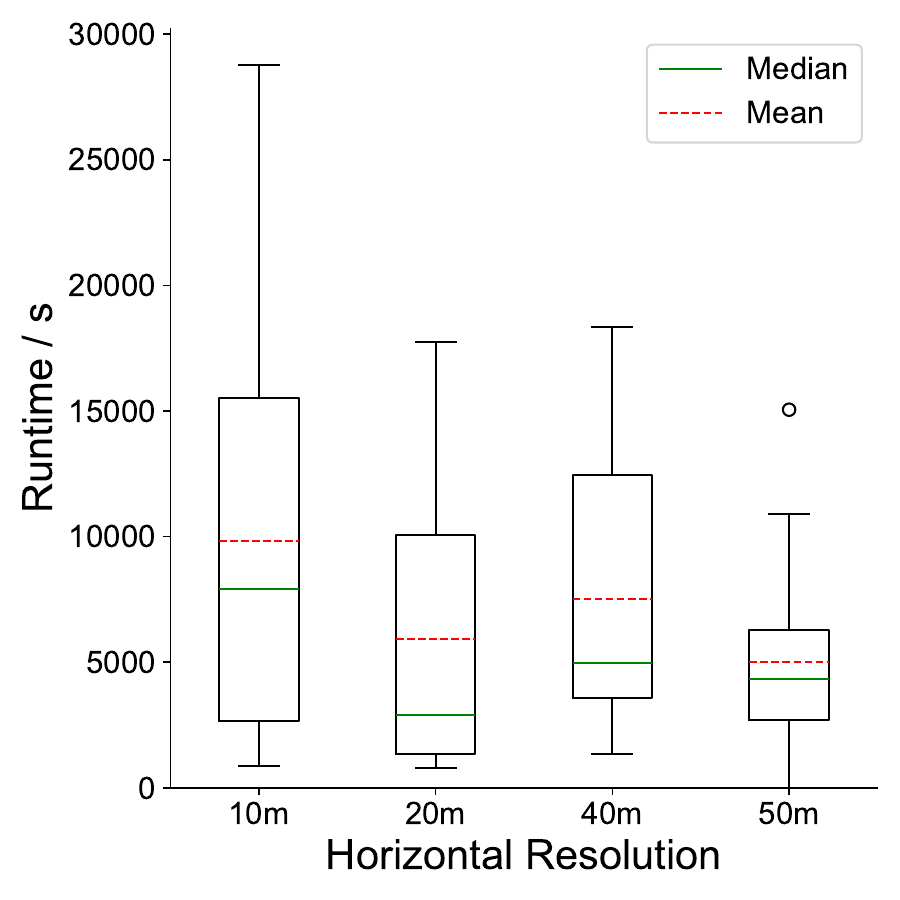}
        \caption{Optimization}
    \end{subfigure}
    \caption{
        \textbf{Runtime requirements of compliant and effective UAV routing:}
        Here, we show the runtime of each step in our routing framework, dependent on the horizontal resolution of the search graph (distance between nodes).
        The runtime of VIAS' many-objective optimization dominates the routing process strongly, being orders of magnitude slower than the preprocessing (Dijkstra) and parameterization (StaR Map computation), but being less dependent on the graph's density.
    }
    \label{fig:runtimes}
\end{figure}

In our experiments, we apply our framework to generate two paths for each of the $13 \times 13$ individual airspace volumes across Paris, each spanning $\SI{1}{\kilo\meter\squared}$ of land.
Figure~\ref{fig:runtimes} shows the time needed for each major step in our pipeline, namely the computation of StaR Map parameters, VIAS' preparation of physical objectives, and the Dijkstra-based initial search, and VIAS' evolutionary generation of the Pareto set.

Although the initialization of our framework can be done within minutes, the evolutionary search takes considerably more time and limits the use of many-objective optimization for real-time applications.
Instead, we aim to provide a traffic network of compliant and effective UAV paths in an offline fashion.
Note how, according to Figure~\ref{fig:runtimes}, StaR Maps' runtime is much less dependent on the desired resolution of the search graph.

\begin{figure}
    \begin{subfigure}{0.55\linewidth}
        \includegraphics[width=\linewidth]{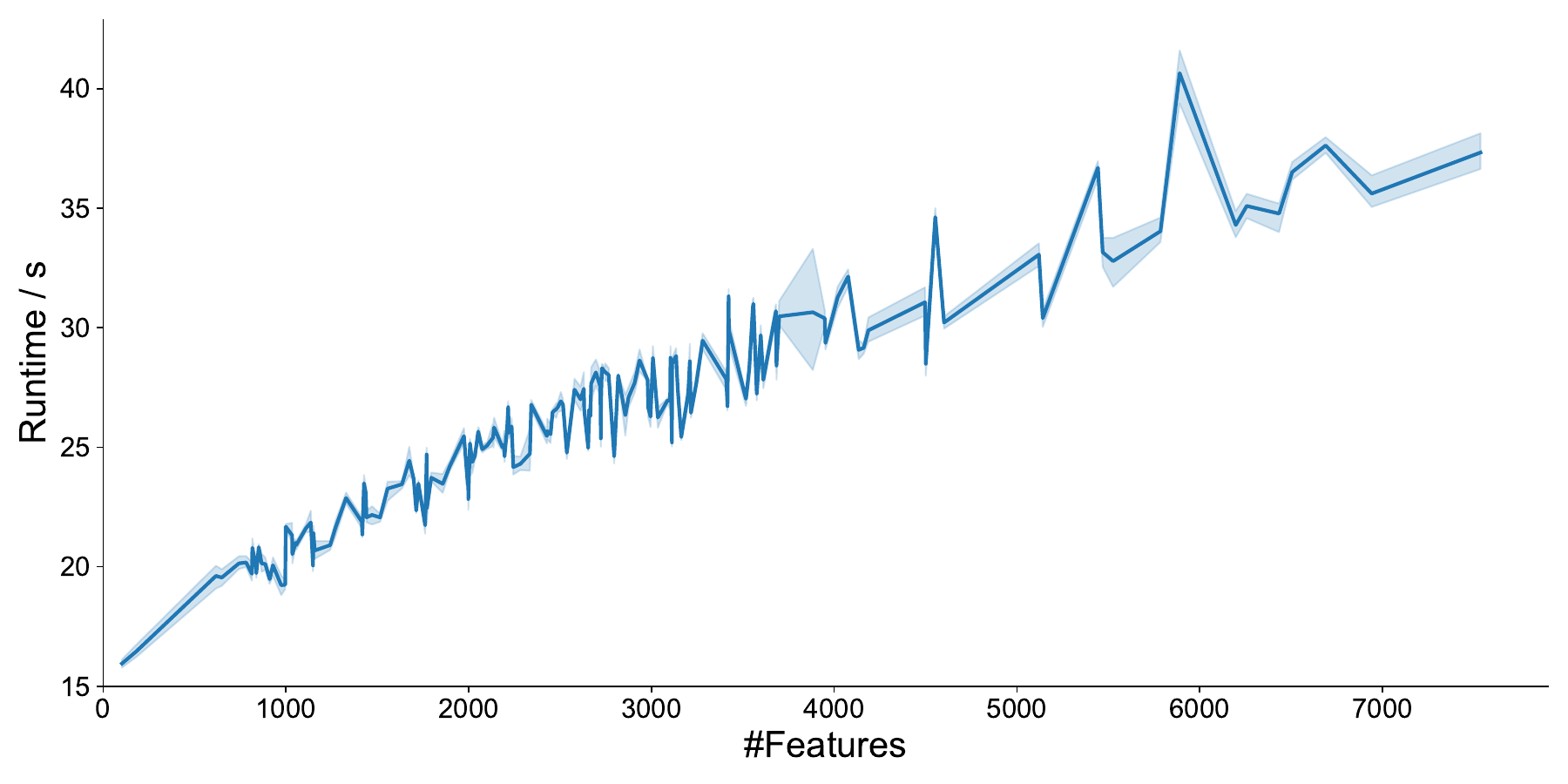}
        \caption{StaR Map Runtime per Feature}
    \end{subfigure}
    \begin{subfigure}{0.375\linewidth}
        \includegraphics[width=\linewidth]{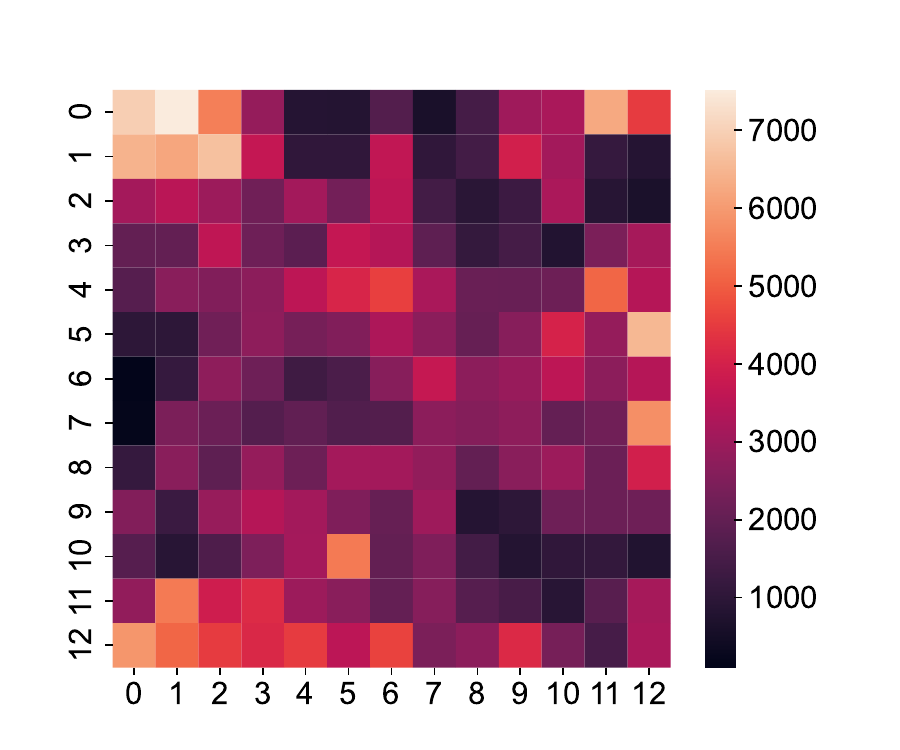}
        \caption{Feature Density across Paris}
    \end{subfigure}
    \caption{
        \textbf{StaR Map computation times depend on feature count:}
        In contrast to VIAS' computation time, StaR Maps' runtime is dominated by the complexity of the environment, i.e., the number of features to consider during sampling of the probabilistic spatial relation (a).
        Hence, it is important to consider the specific environment in which routing will take place, e.g., in our Paris example, the density of features differs greatly across the individual $\SI{1}{\kilo\meter\squared}$ patches used for routing (b).
    }
    \label{fig:star_map_runtimes_per_feature}
\end{figure}

Instead, as shown in Figure~\ref{fig:star_map_runtimes_per_feature}, StaR Maps becomes slower with a growing number of map features to consider in the sampling process.
Hence, depending on the complexity of the application area, e.g., the number of buildings or roads, StaR Maps computation time may grow considerably.

    \section{Conclusion}
\label{sec:conclusion}

\subsection{Summary}

We have presented a compliant and effective routing for UAVs through Hybrid Multi-Objective Optimization within Probabilistic Mission Design (ProMis).
As such, our framework enables UAV operators to jointly encode physical objectives for effectiveness and airspace regulations for compliance. 
While the former is modeled as cost functions over UAV paths or points in navigation space, the latter is facilitated through probabilistic first-order logic and spatial reasoning parameterized by Statistical Relational Maps (StaR Map).

As an integral part of our framework, we have presented the Constitutional Language (CoLa) as a unified description language to formalize an agent's navigational objectives and constraints in a single model.
To this end, CoLa lets the user define high-level mission parameters such as the aviation license of the employed operator or the time of day at which the mission is conducted.
These parameters hereby span up a space for Explanation and Optimization steps beyond path optimization, i.e., showing which part of the setting leads to a denial of Clearance for the intended path or what setting would yield ideal circumstances from a legal viewpoint.
Furthermore, CoLa allows for encoding aviation rules across a StaR Map parameterized space as scalar fields of identically and independently distributed probabilities of satisfied constraints under uncertainty.

We have shown how Hybrid Multi-Objective Optimization in ProMis enables large-scale, physically and legally informed UAV routing in a statistical evaluation of the framework's application in Paris, France.
Based on low-cost, crowd-sourced map data provided by OpenStreetMap, a set of physical objectives about the agent's path and states, as well as a simplified model of legal and safety constraints, our experiments demonstrate how these methods can create networks of compliant paths over vast human-inhabited spaces.
Along these lines, we have illustrated how this setting allows for a meta-analysis of the satisfiable airspace, i.e., how to decide Clearance for individual paths, Explain the impact of mission parameters, and Optimize for the most suitable ones with regard to compliance.
In our experiments, this reduces the path rejection area over curve, e.g., by approximately $80\%$ in the case of knee point paths.

\subsection{Limitations and Future Work}

Probabilistic inference over legal requirements and operational preferences is a computationally demanding endeavor.
Not only does the point-wise execution of the reasoning pipeline not come free of charge, but the granularity at which the navigation space is sampled may quickly exceed what is practical when quick decisions are required.
Hence, the presented methods in their current form do not trivially allow for real-time routing where parameters to regulatory constraints are frequently updated.
To remedy this issue, smart utilization of memoization and online adaptation of the reasoning mechanisms may lead to the necessary improvements.  

Furthermore, while probabilistic first-order logic and StaR Maps allow for a rich vocabulary of spatial reasoning, higher-order logic or temporal components were not considered in this work.
Such extensions might be necessary to fully capture the restrictions of traffic systems, e.g., when considering public aviation regulations.

Finally, we have merely demonstrated a static set of rules, encapsulating a small excerpt of possible expert knowledge to pass onto the system.
When considering more general robotics applications, it is important to endow the system with the necessary learning capabilities, i.e., extracting weighted rules from experience and connecting the framework to a library of sensors and deep learning models to capture the agent's perception into its reasoning and subsequent decision-making.
As our model of airspace constraints is based on probabilistic first-order logic, end-to-end learning with neural networks is possible, paving the way toward fully leveraging neuro-symbolic capabilities for compliant and effective UAV routing.

\section*{Acknowledgments}
Simon Kohaut and Nikolas Hohmann gratefully acknowledge the financial support from Honda Research Institute Europe (HRI-EU).
The Eindhoven University of Technology authors received support from their Department of Mathematics and Computer Science and the Eindhoven Artificial Intelligence Systems Institute.
Map data \copyright~OpenStreetMap contributors, licensed under the Open Database License (ODbL) and available from \url{https://www.openstreetmap.org}.
Map styles \copyright~Mapbox, licensed under the Creative Commons Attribution 3.0 License (CC BY 3.0) and available from \url{https://github.com/mapbox/mapbox-gl-styles}.
Cell tower data \copyright~OpenCellID, licensed under the Creative Commons Attribution ShareAlike 4.0 License (CC BY-SA 4.0) and available under \url{https://wiki.opencellid.org}.

    \bibliographystyle{unsrtnat}
    \bibliography{literature}
    
\end{document}